\documentclass[sigconf]{acmart}
\usepackage{amsmath,epsfig}
\usepackage{cleveref}

\crefformat{section}{\S#2#1#3} 
\crefformat{subsection}{\S#2#1#3}
\long\def\comment#1{}

\usepackage{makecell}
\usepackage{color}
\usepackage{tikz}
\usepackage{bm}
\usepackage{graphicx}
\usepackage{xspace}
\usepackage{subfigure}
\usepackage{caption}
\usepackage{amsmath}
\usepackage{float}
\usepackage{graphicx}
\usepackage{url}
\usepackage{epsfig}
\usepackage{mathrsfs}
\usepackage{multirow}
\usepackage{color}
\usepackage{epstopdf}
\usepackage{epsf}
\usepackage[ruled,vlined, noend]{algorithm2e}
\usepackage{listings}
\usepackage[nounderscore]{syntax}
\usepackage{setspace}
\usepackage{framed}
\usepackage{hyperref}
\usepackage{mathtools}
\usepackage{physics}
\usepackage{enumitem}
\usepackage{pifont}
\usepackage{fancybox}
\usepackage{colortbl}
\usepackage{bbding}

\makeatletter

\newcommand{\Rmnum}[1]{\expandafter\@slowromancap\romannumeral #1@}
\definecolor{LightCyan}{rgb}{0.88,1,1}
\definecolor{LightRed}{rgb}{1,0.88,1}
\definecolor{LightYellow}{rgb}{1,1,0.88}
\definecolor{LightGray}{gray}{0.8}
\makeatother

\let\oldAA\AA
\renewcommand{\AA}{\text{\normalfont\oldAA}}

\usepackage{balance}

\AtBeginDocument{%
  \providecommand\BibTeX{{%
    \normalfont B\kern-0.5em{\scshape i\kern-0.25em b}\kern-0.8em\TeX}}}

\copyrightyear{2024}
\acmYear{2024}
\setcopyright{acmlicensed}\acmConference[CIKM '24]{Proceedings of the 33rd ACM International Conference on Information and Knowledge Management}{October 21--25, 2024}{Boise, ID, USA} \acmBooktitle{Proceedings of the 33rd ACM International Conference on Information and Knowledge Management (CIKM '24), October 21--25, 2024, Boise, ID, USA}
\acmDOI{10.1145/3627673.3679529} \acmISBN{979-8-4007-0436-9/24/10}

\newcounter{remark}[section]
\renewcommand{\theremark}{\nthesection.\arabic{remark}}

\newcommand{\stitle}[1]{\vspace{1ex} \noindent{\bf #1}}

\newcommand{\kw}[1]{{\ensuremath {\mathsf{#1}}}\xspace}

\newcommand{\beqn}{\begin{eqnarray*}}
\newcommand{\eeqn}{\end{eqnarray*}}

\newcounter{ccc}

\newcommand{\kwnospace}[1]{{\ensuremath {\mathsf{#1}}}}

\newcommand{\softmax}{\mathsf{softmax}}

\newcommand{\Real}{\mathbb{R}}

\newcommand{\LR} {{\sl LR}\xspace}
\newcommand{\ECC} {{\sl ECC}\xspace}

\newcommand{\SciBERT} {{\sl SciBERT}\xspace}
\newcommand{\PubBERT} {{\sl PubMedBERT}\xspace}
\newcommand{\BioBERT} {{\sl BioBERT}\xspace}
\newcommand{\BleuBERT} {{\sl BlueBERT}\xspace}
 
\newcommand{\LEAP} {{\sl LEAP}\xspace}
\newcommand{\RETAIN} {{\sl RETAIN}\xspace}
\newcommand{\GAMENet} {{\sl GAMENet}\xspace}

\newcommand{\MICRON} {{\sl MICRON}\xspace}
\newcommand{\SafeDrug} {{\sl SafeDrug}\xspace}
\newcommand{\COGNet} {{\sl COGNet}\xspace}
\newcommand{\DrugRec} {{\sl DrugRec}\xspace}
\newcommand{\MolRec} {{\sl MoleRec}\xspace}

\newcommand{\ClinicalBERT} {{\sl ClinicalBERT}\xspace}
\newcommand{\Encoder}{\kwnospace{PLM}\mbox{-}\kw{Encoder}}

\newcommand{\NLACMR} {{\sl NLA-MMR}\xspace}

\newcommand{\mimicIII}{\kwnospace{MIMIC}\mbox{-}\kw{III}}
\newcommand{\mimicIV}{\kwnospace{MIMIC}\mbox{-}\kw{IV}}
\newcommand{\eICU}{\kw{eICU}}

\begin{document}

\title{Natural Language-Assisted Multi-modal Medication Recommendation}



\author{Jie Tan}
\email{jtan@se.cuhk.edu.hk}
\affiliation{%
  \institution{The Chinese University of Hong Kong}
  \city{Hong Kong}
  \country{China}
}

  \author{Yu Rong}
\email{yu.rong@hotmail.com}
\authornote{Co-corresponding authors.}
\affiliation{%
  \institution{DAMO Academy, Alibaba Group\\ Hupan Lab}
  \city{Hangzhou}
  \country{China}}

\author{Kangfei Zhao}
\email{zkf1105@gmail.com}
\affiliation{%
  \institution{Beijing Institute of Technology}
  \city{Beijing}
  \country{China}
}
\authornotemark[1]

\author{Tian Bian}
\email{tbian@se.cuhk.edu.hk}
\affiliation{%
  \institution{The Chinese University of Hong Kong}
  \city{Hong Kong}
  \country{China}
}

\author{Tingyang Xu}
\email{xuty\_007@hotmail.com}
\affiliation{%
  \institution{DAMO Academy, Alibaba Group\\ Hupan Lab}
  \city{Hangzhou}
  \country{China}}

\author{Junzhou Huang}
\email{jzhuang@uta.edu}
\affiliation{%
  \institution{University of Texas at Arlington}
  \city{Arlington, Texas}
  \country{United States}
}

\author{Hong Cheng}
\email{hcheng@se.cuhk.edu.hk}
\affiliation{%
  \institution{The Chinese University of Hong Kong}
  \city{Hong Kong}
  \country{China}
}

\author{Helen Meng}
\email{hmmeng@se.cuhk.edu.hk}
\affiliation{%
  \institution{The Chinese University of Hong Kong}
  \city{Hong Kong}
  \country{China}
}







\renewcommand{\shortauthors}{Jie Tan et al.}





\begin{abstract}
Combinatorial medication recommendation~(CMR) is a fundamental task of healthcare, which offers opportunities for clinical physicians to provide more precise prescriptions for patients with intricate health conditions, particularly in the scenarios of long-term medical care.
Previous research efforts have sought to extract meaningful information from electronic health records (EHRs) to facilitate combinatorial medication recommendations. Existing learning-based approaches further consider the chemical structures of medications, but ignore the textual medication descriptions in which the functionalities are clearly described.
Furthermore, the textual knowledge derived from the EHRs of patients remains largely underutilized. 
To address these issues, we introduce the Natural Language-Assisted Multi-modal Medication Recommendation~(\NLACMR), a multi-modal alignment framework designed to learn knowledge from the patient view and medication view jointly. Specifically, \NLACMR formulates CMR as an alignment problem from patient and medication modalities. 
In this vein, we employ pretrained language models (PLMs) to extract in-domain knowledge regarding patients and medications, serving as the foundational representation for both modalities.
In the medication modality, we exploit both chemical structures and textual descriptions to create medication representations. In the patient modality, we generate the patient representations based on textual descriptions of diagnosis, procedure, and symptom. 
Extensive experiments conducted on three publicly accessible datasets demonstrate that \NLACMR achieves new state-of-the-art performance, with a notable average improvement of 4.72\% in Jaccard score. Our source code is publicly available on \url{https://github.com/jtan1102/NLA-MMR_CIKM_2024}.
\end{abstract}

\maketitle

\section{Introduction}
\label{sec.intro}
The implementation of digital patient records in healthcare institutions has boosted the development of a substantial repository of information known as Electronic Health Records (EHRs). This valuable asset holds significant potential for a variety of applications within the medical field, such as predicting mortality rates, forecasting treatment outcomes, and offering medication recommendations. As one of the fundamental tasks of healthcare, combinatorial medication recommendation (CMR) has attracted a lot of attention in recent years, and achieved significant progress in providing personalized and safe medication recommendations for patients. This advancement is especially beneficial for the elderly who have been suffering from chronic illnesses for a prolonged time \cite{ruksakulpiwat2023does,shoenbill2023artificial}.

\begin{figure}[t]
\begin{center}
\includegraphics[width=1\columnwidth]{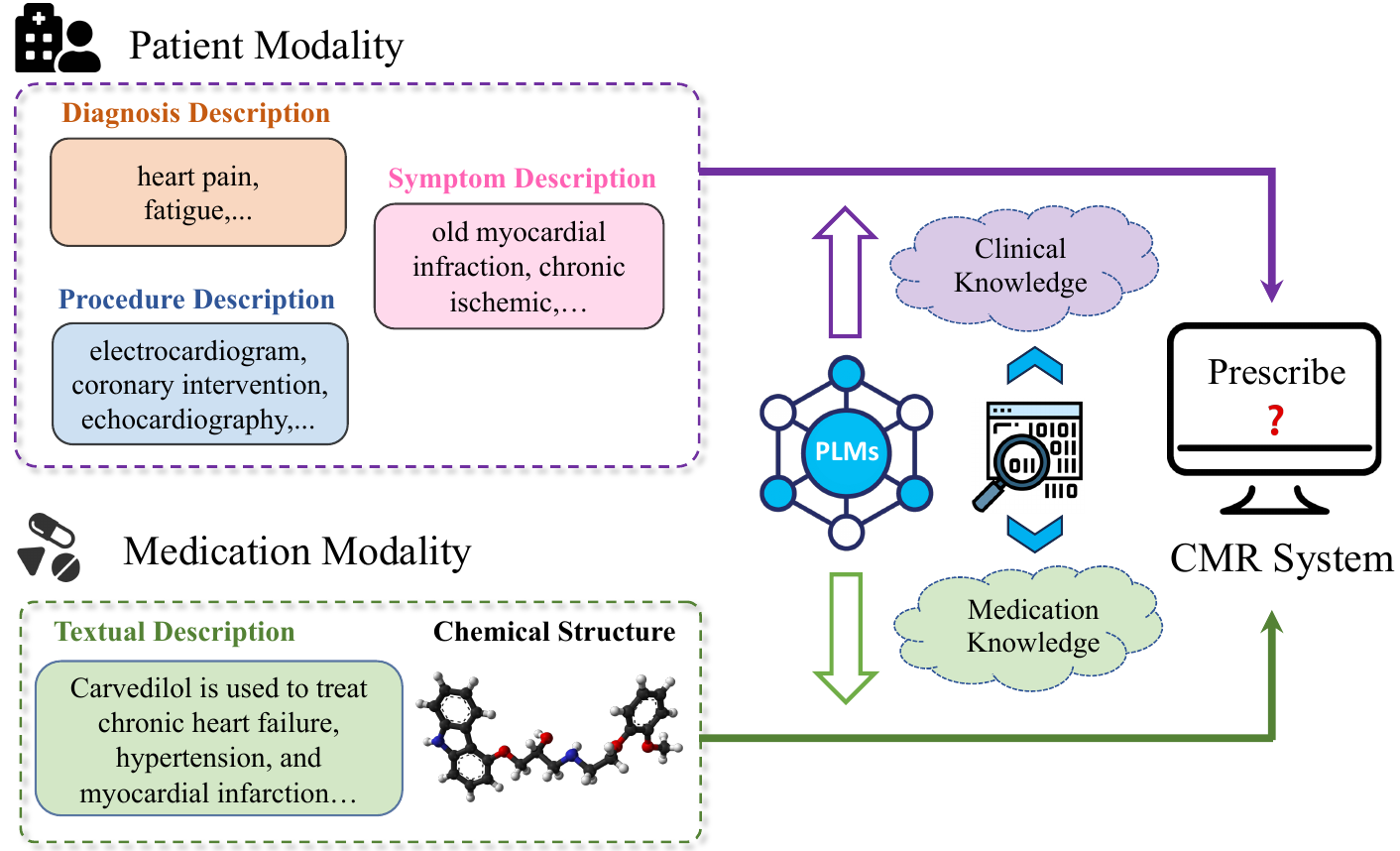}
\end{center}
\vspace*{-0.1cm}
\caption{Illustration of combinatorial medication recommendation with help of knowledge extraction from PLMs
}
\label{fig:motivation}
\end{figure}

A multitude of deep learning models have been
designed to solve the CMR task in the literature, which not only use the EHRs (e.g., diagnoses, procedures, and symptoms), but also leverage the chemical structures of drugs. 
For example, \SafeDrug~\cite{DBLP:conf/ijcai/YangXMGS21} developed dual molecular graph encoders to embed global and local molecular structures. \MolRec~\cite{DBLP:conf/www/YangZWY23} proposed a molecular substructure-aware attentive method for CMR. 
\DrugRec~\cite{DBLP:conf/nips/Sun0LCW022} proposed to represent drugs with the SMILES, a line-based representation (i.e., string) of molecules.
Besides EHRs and chemical structures, we observe there are a vast amount of textual descriptions of drug molecules that are human-understandable and easily accessible on platforms such as PubChem~\cite{DBLP:journals/nar/00020CGHHLSTYZ021} and DrugBank~\cite{DBLP:journals/nar/WishartFGLMGSJL18}. 
These texts describe the functionalities of drug molecules and reveal their therapeutic purposes. Unfortunately, none of the existing methods try to exploit the textual medication descriptions, thus failing to establish the semantic relations between the textual descriptions of patients and medications.
Consider the example depicted in Figure~\ref{fig:motivation} with two modalities of patient and medication, respectively. In the patient modality, an EHR shows that the patient with the symptoms \textit{``heart pain''} and \textit{``fatigue''} is diagnosed with diseases \textit{``old myocardial infraction''} and \textit{``chronic ischemic''}, and then is prescribed with a set of drugs \textit{``carvedilol''}, \textit{``aspirin''} and \textit{``acetaminophen''}.  In the medication modality, the textual medication description reveals that 
\textit{``Carvedilol is used to treat chronic heart failure,
hypertension, and
myocardial infarction''}.
This example illustrates that the textual medication description can supplement the chemical structures to enhance the expressiveness of medication representations, and help establish the semantic relations between the patients and medications. This motivates us to leverage the useful textual medication description and consider patient and medication as two distinct modalities for medication recommendation.

In another direction, recent studies have attempted to incorporate structured domain knowledge, such as biomedical knowledge graphs (BioKG)~\cite{DBLP:conf/aaai/ShangXMLS19,DBLP:conf/www/WuQJQW22} and drug-drug interactions (DDI)~\cite{DBLP:conf/ijcai/YangXMGS21, DBLP:conf/nips/Sun0LCW022, DBLP:conf/icde/BianJLXRSKM023} to enhance the CMR performance. However, the quality of such structured domain knowledge may degrade due to potential information loss and biased sampling when it is extracted from raw medical documents. A more effective approach is to employ the pretrained language models (PLMs), which are pretrained on web-scale textual data from the chemical and biological domain to capture clinical specialist knowledge in raw texts for medication recommendation. Representations learned by PLMs like~\PubBERT~\cite{DBLP:journals/health/GuTCLULNGP22} and~\ClinicalBERT~\cite{lee2020clinical} achieve strong performance across many domain-specific applications.
Intuitively, using these PLMs to encode the textual knowledge of the patient and medication modalities can provide additional gains over conventional encoding methods in the CMR task because they effectively fuse various types of information, such as diagnoses, procedures, symptoms and medications in a unified global semantic space. 

In light of these observations, we propose a novel model, called Natural Language-Assisted Multi-modal Medication Recommendation~(\NLACMR), to make medication recommendations by effectively harnessing the wealth of clinical knowledge in the EHRs and drugs. \NLACMR incorporates the textual medication descriptions for molecular representation learning. It also integrates expert knowledge by leveraging PLMs to capture various types of semantic relations between the texts in the EHRs and medication descriptions. Specifically, \NLACMR aligns patient and medication modalities utilizing a cross-modal alignment module based on the patient's prescriptions in the EHRs. In the prediction phase, \NLACMR transforms a recommendation problem into a similarity retrieval problem that aims to find the most similar medication representation given the patient's representation. Our contributions in this paper are summarized as follows:
\begin{itemize}[leftmargin=*]
    \vspace{-0.2cm}
    \item To the best of our knowledge, we are the first to consider the patient and medication as two distinct modalities and design a multi-modal alignment framework \NLACMR to learn their representations in a unified latent space. \NLACMR converts the recommendation task into a similarity retrieval problem, enabling better generalizability and transferability of the model.
    \item We propose utilizing the PLMs as a fundamental building block to construct the representations of the patient and medication modalities by extracting expert knowledge from their descriptions. For the medication modality, we incorporate both chemical structures and medication descriptions for the medication representation learning. 
    \item 
    Experimental results on three publicly available healthcare datasets demonstrate that our proposed method is highly effective compared to the state-of-the-art methods. Our model boosts Jaccard by 2.86\%, 7.01\%, 4.29\% on \mimicIII, \mimicIV, and \eICU datasets, respectively. 
\end{itemize}

\vspace{-0.2cm}
\section{Preliminaries}
\label{sec:pre}

For a patient $v$, his/her electrical health records (EHRs) are represented as a sequence $R_{v} = [x_{v}^{(1)}, x_{v}^{(2)}, \cdots, x_{v}^{(T_{v})}]$, where $x_{v}^{(t)}$ represents the health record from the $t$-th clinical visit and $T_{v}$ is the total number of visits of the patient $v$. When the context is clear, we omit the subscript $v$ to simplify notations. Then, the $t$-th visit in the EHR of a patient can be represented by a  $x^{(t)} = (d^{(t)}, p^{(t)}, s^{(t)}, m^{(t)})$, where  $d^{(t)} \in \{0,1\}^{|\mathcal{D}|}$, $p^{(t)} \in \{0,1\}^{|\mathcal{P}|}$,  $s^{(t)} \in \{0,1\}^{|\mathcal{S}|}$, $m^{(t)} \in \{0,1\}^{\mathcal{{|{M}|}}}$ are multi-hot diagnosis, procedure, symptom, and medication vectors, respectively. $\mathcal{D}$, $\mathcal{P}$, $\mathcal{S}$, $\mathcal{M}$ indicate the set of possible diagnoses, procedures, symptoms and medications, respectively, and $|\cdot|$ denotes the cardinality of a set. 
As shown in Figure~\ref{fig:motivation}, our approach treats the patient and medication as two modalities to consider their information.

\noindent\textbf{Patient Modality.}
Following~\cite{DBLP:conf/nips/Sun0LCW022}, we model the patient modality as a series of clinical visits containing the  diagnosis, procedure, or symptom, which reflect the health state of this patient. Thus, the $t$-th visit for the patient $v$ is defined as $r_v^{(t)} = (r_d^{(t)}, r_p^{(t)}, r_s^{(t)})$, where $r_d^{(t)}$, $r_p^{(t)}$, and $r_s^{(t)}$
represent the textual description of diagnoses in $d^{(t)}$, procedures in  $p^{(t)}$, and symptoms in $s^{(t)}$, respectively. Therefore, the patient modality of $v$ is denoted as $v = [r_v^{(t)}]_{t=1}^{T_v}$.


\noindent\textbf{Medication Modality.}
To enrich the medication modality, we collect the textual description $r$ for each molecule in $\mathcal{M}$, which provides a high-level explanation of the molecule's functionality.
Additionally,
for each medication in $\mathcal{M}$, we represent its molecular structure by an undirected graph $g = (\mathcal{A}, \mathcal{B})$. Here, $\mathcal{A} = \{a_1, \cdots, a_n \}$ is the set of atoms and $\mathcal{B} \subseteq \mathcal{A} \times \mathcal{A}$ is the set of bonds. 
The neighborhood of atom $a_i$ is denoted as $\mathcal{N}(a_i) = \{ a_j | (a_i, a_j) \in \mathcal{B}\}$. 
Therefore, the medication modality is represented as $(r, g)$. 
Figure~\ref{fig:motivation} depicts an example of medication ``Carvedilol''.


\noindent\textbf{Combinatorial Medication Recommendation.}
Given the EHR of a patient till time $t-1$, denoted as $[x^{(1)}, x^{(2)},\cdots, x ^{(t-1)}]$, containing the textual description of the $t$-th diagnosis, procedure, and symptom sets, denoted as $r_d^{(t)}$  $r_p^{(t)}$ and $r_s^{(t)}$, respectively, and a candidate medication set $\mathcal{M}$, the CMR task aims to predict an appropriate combination of medications $\mathcal{M}_{t} \subseteq \mathcal{M}$ for the patient's $t$-th visit using a neural network model $f(\cdot)$. 

\begin{figure*}[t]
\begin{center}
\includegraphics[width=1.93\columnwidth]{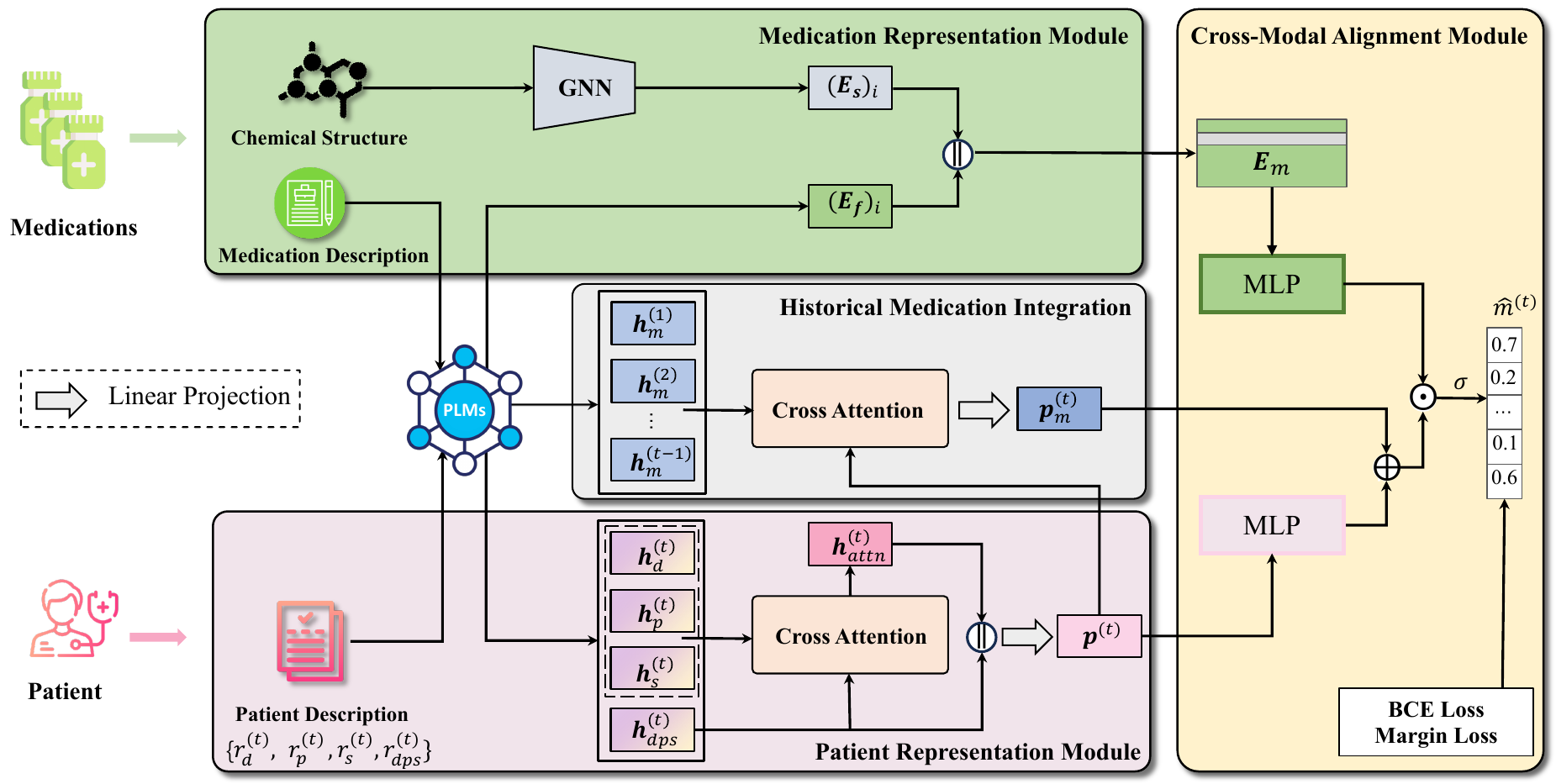}
\end{center}
\vspace*{-0.2cm}
\caption{ The architecture of \NLACMR. \NLACMR is composed of three modules: (a) Patient Representation Module employs PLMs as the base encoder to construct the representation of patient modality from the diagnosis, procedure, and symptom. (b) Medication Representation Module incorporates the embedding derived from textual medication descriptions and chemical structures to build the representation of medication modality. (c) Cross-Modal Alignment Module takes the representation from patient and medication modality as input and aligns them in the same latent space. 
We further consider historical medication usage information to model the patient's clinical history, which can enhance the representation of patient modality.
}
\label{fig:overview}
\vspace*{-3ex}
\end{figure*} 

\section{Methodology}
\label{sec:NLA-MMR}

\subsection{Overview}
\label{sec:overview}

Figure~\ref{fig:overview} illustrates an overview of our proposed drug recommendation approach \NLACMR, which is composed of three modules:
1) the \textbf{patient representation module} using a PLM to encode the texts of diagnosis, procedure, and symptom;
2) the \textbf{medication representation module} generating medication representations by employing the GNN model and a PLM for incorporating molecular structure-level information and textual knowledge, respectively;
3) the \textbf{cross-modal alignment module} responsible for making prescriptions by using two projection layers to calculate the similarity between the patient representation and medication representations on the joint latent space.
In particular, in the patient representation module, 
we conduct cross-attention fusion to fully leverage the different types of textual information present in the patient modality. 
Besides, we integrate the medication information of historical visits to capture dependencies between medication usage in previous visits and the current visit.

\vspace{-0.3cm}
\subsection{Patient Representation Module}
In this section, we construct the patient representation with the textual components defined in the patient modality. 
By leveraging the power of PLMs, we are able to extract valuable expert knowledge from these unprocessed texts, thereby contributing to the development of an advanced medication recommendation model.
Specifically, we first consider the PLM-based representation of the concatenated text of diagnosis, procedure, and symptom as the basic patient representation. Then, we devise a cross-attention fusion mechanism to merge information from the three types of patient descriptions to obtain the final patient representation.

\noindent\textbf{PLM-based Representation.}
We employ a PLM encoder~\cite{DBLP:journals/bioinformatics/LeeYKKKSK20} to extract the representation for the patient modality. 
Give a patient's $t$-th visit $r^{(t)}=(r_d^{(t)}, r_p^{(t)}, r_s^{(t)})$, we concatenate these descriptions into a synthesis text as ${r}_{dps}^{(t)}$ and employ the PLM encoder to obtain its representation: 
\vspace{-0.5ex}
\begin{equation}
\vspace{-0.5ex}
\bm{h}_{dps}^{(t)} = {\Encoder}\left({r}_{dps}^{(t)}\right),
\label{eq:dps_feature}
\vspace{-0.1cm}
\end{equation}
where $\bm{h}_{dps}^{(t)} \in \Real^{d_{enc}}$ is the output of the PLM encoder, which is obtained via an average pooling of the embeddings of all tokens.
Meanwhile, we individually encode the textual descriptions of diagnosis, procedure, and symptom with the PLM and obtain $\bm{h}_d^{(t)}$, $\bm{h}_p^{(t)}$, and~$\bm{h}_s^{(t)}$, which are representations of the corresponding type of textual description in the patient modality.

\noindent\textbf{Cross-Attention Fusion.}
To enhance the basic patient representation in Eq.~\eqref{eq:dps_feature}, we aggregate the features derived from three aspects, $\bm{h}_d^{(t)}$, $\bm{h}_p^{(t)}$, and~$\bm{h}_s^{(t)}$, by utilizing a cross-attention mechanism, weighted by a set of learnable weights $w_c \in \mathbb{R}^{|C|}$, as $\bm{h}_{attn}^{(t)} = \sum_{c \in C} w_c \bm{h}_c^{(t)}, C = \{s,d,p\}$.
Let $\bm{H}^{(t)} \in \Real^{|C| \times d_{enc}}$ be the matrix stacked by the embeddings in $\{\bm{h}_d^{(t)}, \bm{h}_p^{(t)}, \bm{h}_s^{(t)}\}$.
In Eq.~(\ref{eq:pool:att:trans}), 
$\bm{H}_{1} \in \mathbb{R}^{|C| \times d_{K}}$, $\bm{H}_{2} \in \mathbb{R}^{1 \times d_{K}}$are transformed by linear weight matrices $\bm{W}_{1} $, $ \bm{W}_{2} \in \mathbb{R}^{d_{enc} \times d_{K}}$, respectively. 
$w_c$ is computed by the multiplication of $\bm{H}_{1}$ and the transpose of $\bm{H}_{2}$ followed by a $\softmax$ function that normalizes the weights, as Eq.~(\ref{eq:pool:att:weight}) shows.
\vspace{-1ex}
\begin{align}
\vspace{-1ex}
 \label{eq:pool:att:trans}
 \bm{H}_{1} &= \bm{H}^{(t)} \bm{W}_{1},~ \bm{H}_{2} = \bm{h}^{(t)}_{dps} \bm{W}_{2},~\\
 \label{eq:pool:att:weight}
 \{w_c \}_{c \in C} &= \softmax \biggl(\frac{ \bm{H}_{1} \bm{H}_{2}^T}{\sqrt{d_{K}}} \biggr).
\end{align}

Next, we concatenate the aggregated representation $\bm{h}_{attn}^{(t)} \in \Real^{d_{enc}}$ and the basic patient feature $\bm{h}_{dps}^{(t)} \in \Real^{d_{enc}}$ as the patient representation, which can be formulated as:
\vspace{-1ex}
\begin{equation}
\vspace{-1ex}
\label{eq:sdp_attn}
     \bm{p}^{(t)} =  
  \left[\bm{h}_{attn}^{(t)} \| 
  \bm{h}_{dps}^{(t)}\right]\bm{W}_r + \bm{b}_r,  \\
\end{equation}
where $\left[ \| \right]$ denotes the concatenation operation and $\bm{W}_r \in \Real^{2{d_{enc}} \times {d_{enc}}}$ and $\bm{b}_r \in \Real^{d_{enc}}$ are trainable parameters.
\\


\vspace{-0.4cm}
\subsection{Medication Representation Module}

The medication modality encompasses two essential components, namely textual knowledge and structural information. 
As depicted in Figure~\ref{fig:overview}, our module for representing medications is composed of two branches: \emph{the textual description branch} and \emph{the chemical structure branch}. These branches are designed to address the external domain knowledge of molecules through textual descriptions and their intrinsic property  through chemical structures, respectively. To harness the domain-specific knowledge contained within the textual descriptions, we employ a text encoder known as the PLM. Additionally, we employ a GNN model to extract representations at the structure level for drugs based on their chemical structures.
\\
\noindent\textbf{PLM-based Functional Representation.}
The textual description of a drug provides a high-level overview of the molecule’s functionality, which offers valuable insights into the therapeutic potential of the drug and can aid in drug predictions.
In this paper, we extract therapeutic descriptions of the drugs from  DrugBank~\cite{DBLP:journals/nar/WishartFGLMGSJL18}, which illustrates drug effects on the target organism. 
Given the textual medication descriptions, we obtain its PLM-based functional representations by conducting a similar process as in Eq. \eqref{eq:dps_feature}. We collect the textual embeddings of all drugs into a drug matrix $\bm{E}_{f} \in \Real^{\mathcal{|M|}\times d_{enc}}$, where each row corresponds to the textual embedding of a drug. 

\noindent\textbf{Structure-level Representation.}
We incorporate chemical structures to learn the molecule representation.
Specifically, we employ the GNN to model the interactions between all the atoms across the single molecule graph $g$ and obtain the structure-level drug representation. 
The aggregation and combination paradigm is conducted on the graph $g$ of a drug molecule whose nodes are atoms and edges are chemical bonds.
Given the features of the $n$ atoms $\bm{E}^{(k)}_a = \{ \bm{e}^{(k)}_{a_1}, \cdots, \bm{e}^{(k)}_{a_n}\}\in \Real^{n \times d_{s}}$, for each atom $a_i \in \mathcal{A}$, the messages of all the other atoms in the molecule are aggregated to $\bm{e}_i^{(k)} \in \Real^{d_{s}}$(Eq.~(\ref{eq:gin:agg})). Afterward, the aggregated message $\bm{e}_i^{(k)}$ is combined with the feature of the atom $\bm{e}^{(k)}_{a_i}$ by a summation operation, followed by a multilayer perceptron~(MLP), as the combination function of GIN~\cite{DBLP:conf/icml/SatorrasHW21}, generating new atom feature $\bm{e}^{(k + 1)}_{a_i}$ for the next layer~(Eq.~(\ref{eq:gin:combine})). 
\vspace{-1ex}
\begin{align}
\vspace{-1ex}
\bm{e}_i^{(k)} &= \kw{AGG}^{(k)} \left( \left\{ \bm{e}^{(k)}_{a_j} | \forall a_j\in \mathcal{N}(a_i) \right\} \right) \label{eq:gin:agg}, \\
\bm{e}^{(k + 1)}_{a_i} &=  \kw{MLP}^{(k)}\left( \left(1+ \epsilon^{(k+1)} \right) \bm{e}^{(k)}_{a_i} + \bm{e}_i^{(k)} \right), \forall i \in \{1,\cdots,n\},
\label{eq:gin:combine}
\end{align}
where $\epsilon$ can be a learnable parameter or a fixed scalar, it determines the importance of the target node compared to its neighbors, $\bm{e}^{(k)}_{a_j}$ is the feature vector of atom $a_{j}$ at the $k$-th layer, $\mathcal{N}(a_i)$ denotes the neighbors of the atom ${a_i}$. 
\comment{
The initial feature of an atom $h^{(0)}_{a_i}$ is determined by the type of the atom. 
Concretely, let $E_a$ denote the learnable atom embedding table.
}
After applying message passing for $L$ layers, the embeddings of the atoms in the molecule graph are aggregated into a global structural-level  representation as Eq.~\eqref{eq:gin:readout},
\vspace{-1ex}
\begin{align}
\vspace{-1ex}
\bm{e}_{g} &= \kw{Pooling} \left( \left\{ \bm{e}^{(L)}_{a_i} |  \forall i\in \left\{1,\cdots,n \right\} \right\} \right),
\label{eq:gin:readout} 
\end{align}
where the pooling function is the average function.
We employ the same GNN encoder with shared parameters
for each drug molecule (with a total of $\mathcal{|M|}$ different drugs). 
We collect the GNN embeddings of all drugs into a drug memory $\bm{E}_{s} \in \Real^{\mathcal{|M|}\times d_{s}}$, where each row corresponds to the drug's structural representation, $d_{s}$ represents the dimension of the structure-level drug representation.

The textual representation and structural representation are combined to form the final medication embedding matrix $\bm{E}_m \in \Real^{\mathcal{|M|} \times (d_s + d_{enc})}$ as follows:
\vspace{-1ex}
\begin{equation}
\vspace{-1ex}
\label{eq:concat_m}
     \bm{E}_m = 
  [\bm{E}_{s}||
  \bm{E}_{f}].\\
\end{equation}

\comment{DrugBank is by far the most comprehensive database for drug-like molecules, containing detailed information about the drug's names, synonyms, and structure, along with a detailed textual description.}

\subsection{Cross-Modal Alignment Module}
In the cross-modal alignment module, the patient representation $\bm{{p}}^{(t)}$ and medication representations $\bm{E}_{m}$ are fed into two MLPs which map the representations extracted from different modalities to a joint latent space as Eq.~\eqref{eq:crossmodal:mlp}:
\vspace{-1ex}
\begin{align}
\vspace{-1ex}
\label{eq:crossmodal:mlp}
\bm{\Tilde{p}}^{(t)} = \kw{MLP}({\bm{p}}^{(t)}),~
\bm{\Tilde{E}}_m = \kw{MLP}\left(\bm{E}_m\right).
\end{align}
Subsequently, we use inner product operation $\odot$ to compute the similarity of the patient feature $\bm{\Tilde{p}}^{(t)}$, and the medication features $\bm{\Tilde{E}}_{m}$ in the joint space as Eq.~\eqref{eq:crossmodal:predict}:
\begin{align}
\label{eq:crossmodal:predict}
\hat{m}_i^{(t)} = \sigma \left(\bm{\Tilde{p}}^{(t)} \odot   \left(\bm{\Tilde{E}}_m\right)_{i}  \right), \forall i \in \{1,\cdots,{|\mathcal{M}|}\}.
\end{align}
The predicted recommendation probabilities $\hat{m}^{(t)} \in [0, 1]^{|\mathcal{M}|}$ are obtained by a sigmoid function $\sigma$.

\comment{However, previous approaches have often overlooked the explicit modeling of medication history representations, focusing primarily on the current visit representation as the final patient representation to capture longitudinal dependencies and predict medications.
We next generate a safe drug recommendation by comprehensively modeling the drug molecule databases.}

\vspace{-0.2cm}
\subsection{Historical Information Integration}
Within the realm of medication recommendation, it is of utmost importance to effectively model the historical clinical information of patients, particularly those suffering from chronic ailments that necessitate long-term medication. Prior studies commonly employ the extraction of historical features, such as diagnosis, procedure, and symptom, to capture the patient's clinical trajectory~\cite{DBLP:conf/www/YangZWY23, DBLP:conf/ijcai/YangXMGS21}. Nevertheless, the observed patient data often suffers from incompleteness and inadequacy, subsequently resulting in imprecise predictions in the CMR task~\cite{DBLP:conf/nips/Sun0LCW022}. To overcome this limitation, we propose a historical information integration mechanism that directly assimilates the patient's medication history. This module generates a comprehensive representation of the patient's historical clinical condition by duly considering the temporal patterns of medication usage. By leveraging this integrated information, we aim to enhance the accuracy and precision of medication recommendations.

Firstly, we denote $r_m^{(t)}$ as the textual description of the medications used in the $t$-th visit
and obtain the medication representations of all the past visits $\bm{h}^{(i)}_m \in \Real^{d_{enc}}$ for the visit $i \in \{ 1, \cdots, t-1\}$
through the similar process as in Eq.~\eqref{eq:dps_feature}:
\vspace{-1ex}
\begin{equation}
\vspace{-1ex}
\label{eq:med_sequence}
\bm{h}_{m}^{(i)} = {\Encoder\left({r}_{m}^{(i)}\right)},~\forall i \in \{1, \cdots, t-1 \}.
\end{equation}

Then, we use the patient representation in Eq.~\eqref{eq:sdp_attn}, $\bm{p}^{(t)}$, as the query vector to calculate the selection score $\{ w^{(i)}_m | i \in {1, \cdots, t-1}\}$ along the historical medication combination features. 
Let $\bm{H}_{m} \in \Real^{(t-1) \times d_{enc}}$ be the matrix stacked by the historical medication embeddings $\{\bm{h}_m^{(i)} | i \in \{1, \cdots, t-1 \}\}$, we calculate $w_m$ as follows:
\begin{align}
 \label{eq:med:att}
 \bm{H}_{3} &= \bm{H}_{m} \bm{W}_{3},~ \bm{H}_{4} = \bm{p}^{(t)} \bm{W}_{4},~\\
 \label{eq:med:att:weight}
 \{w_m^{(i)} \}_{i \in \{1,\cdots,t-1\}} &= \softmax \biggl(\frac{ \bm{H}_{3} \bm{H}_{4}^T}{\sqrt{d_{K}}} \biggr),
\end{align}

\noindent where $\bm{H}_{3} \in \Real^{(t-1) \times d_{K}}$, $\bm{H}_{4} \in \Real^{1 \times d_{K}}$, and $\bm{W}_{3} , \bm{W}_{4} \in \mathbb{R}^{d_{enc} \times d_{K}}$ are learnable parameters. Finally, the historical medication representation of the patient can be computed as follows:
\vspace{-1ex}
\begin{align}
\vspace{-1ex}
\bm{{p}}_{m}^{(t)} = \left(\sum_{i \in \{1, \cdots, t-1\}} w^{(i)}_m \bm{h}_{m}^{(i)}\right)\bm{W}_m +\bm{b}_m ,
\label{eq:med:att:pm}
\end{align}

\noindent where $\bm{W}_m \in \mathbb{R}^{d_{enc} \times d_{K}}$ and $\bm{b}_m \in \mathbb{R}^{d_{K}}$ are trainable parameters. The embedding $\bm{\Tilde{p}}^{(t)}$ + $\bm{{p}}_{m}^{(t)}$ can be considered as the patient feature for the alignment in Eq.~\eqref{eq:crossmodal:predict}. 


\subsection{Training Objective}  
For each patient in the training set, we use the bce loss in Eq.~\eqref{eq:loss:bce} as the optimization objective, which is a summation for all the $T$ times visits. 
Here, $m^{(t)} \in \{0, 1\}^{|\mathcal{M}|}$ and $\hat{m}^{(t)} \in [0, 1]^{|\mathcal{M}|}$ are the ground truth label and recommendation probabilities for a drug set $\mathcal{M}$, respectively. 
\vspace{-0.5ex}
\begin{equation}
\label{eq:loss:bce}
\mathcal{L}_{{bce}} = -\sum_{t = 1}^{T} \sum_{i = 1}^{|\mathcal{M}|} \left(m_{i}^{(t)} \log \hat{m}_{i}^{(t)} + (1-m_{i}^{(t)})\log (1 - \hat{m}_{i}^{(t)})\right)
\end{equation}

We utilize the margin loss $\mathcal{L}_{{mar}}$ in Eq.~\eqref{eq:loss:multi}, aiming to enlarge the discrepancy between the predicted probability of 
positive and negative cases in the prescription.
\vspace{-0.5ex}
\begin{equation}
\begin{aligned}
\label{eq:loss:multi}
\vspace{-1ex}
\mathcal{L}_{{mar}} = \sum_{t=1}^{T}
\sum_{\{i|m_{i}^{(t)}=1\}} 
\sum_{\{j|m_{j}^{(t)}=0\}}
\frac{\max \{1-(\hat{m}_{i}^{(t)}-\hat{m}_{j}^{(t)}),0 \}}{|\mathcal{M}|}.
\end{aligned}
\end{equation}

The final training objective of the recommendation prediction can be formulated as:
\vspace{-0.5ex}
\begin{equation}
\begin{aligned}
\label{eq:final_loss}
\mathcal{L} = \alpha \mathcal{L}_{{bce}} + (1-\alpha){\mathcal{L}_{{mar}}},
\end{aligned}
\end{equation}
where $\alpha$ is a hyperparameter to balance these two loss terms.

\vspace{-0.2cm}
\subsection{Prediction of $\mathcal{M}_{t}$}
In the prediction stage, given the medication set $\mathcal{M}$ and the model $f(\cdot)$, we first obtain the features for all the medications as $\bm{\Tilde{E}}_{m}$. For the $t$-th visit of the patient $v$, we acquire its representation $\bm{p}_{v} = \bm{\Tilde{p}}^{(t)}+\bm{{p}}_{m}^{(t)}$, by employing $f(\cdot)$ as the query embedding. 
Subsequently, we compute the similarities between $\bm{p}_{v}$ and all medications in $\bm{\Tilde{E}_{m}}$ and then sort the medications based on similarity in ascending order. To produce the final recommendation set $\mathcal{M}_{t}$, we apply the predefined threshold value $\delta$ to select medications with a similarity greater than $\delta$.
\vspace{-0.2cm}
\subsection{Complexity Analysis}
Given $N$ patients, $|\mathcal{M}|$ medications, $d_{K}$ as the hidden dimension of MLP layer, $d_{enc}$ as the output dimension of PLMs, $d_{s}$ as the output dimension of GNN encoder, $T$ as the maximum number of historical visits for all patients, and $iter$ as the number of iterations, we analyze the complexity of \NLACMR as follows.
\\
\textbf{Time complexity}. The time complexity of linear feature transformation (Eq.(\ref{eq:sdp_attn})) is $O(T\cdot d_{enc}^{2})$, and that of the attention between patient textual descriptions (Eq.(\ref{eq:pool:att:trans})-(\ref{eq:pool:att:weight}))
is $O(T\cdot d_{enc}\cdot d_{K})$. 
For the GNN encoder (Eq.(\ref{eq:gin:agg})-(\ref{eq:gin:readout})), the time complexity is $O(|\mathcal{M}|\cdot L\cdot (e_{0}\cdot d_{s}+n\cdot d_{s}^2)$, where $L$ is the number
of GNN layers, $e_{0}$ is the number of edges in a molecular graph, and $n$ denotes
the number of atoms. 
The time complexity of MLP layers in Eq.(\ref{eq:crossmodal:mlp}) is $O(T\cdot d_{enc}\cdot d_{K})$ for patients and $O(|\mathcal{M}|\cdot(d_{enc}+d_s)\cdot d_{K})$ for medications, respectively. The complexity of the output layer in Eq.(\ref{eq:crossmodal:predict}) for medication prediction is $O(T\cdot |\mathcal{M}|\cdot d_{K})$. 
The complexity of attention between all past visits (Eq.(\ref{eq:med:att})-(\ref{eq:med:att:pm})) is $O(T\cdot d_{enc}\cdot d_{K})$. 
With an average degree of 2.1 in molecular graphs for \mimicIII and \mimicIV, and $d_{s} < d_{K} < d_{enc}$,
the total time complexity is $O(iter\cdot N\cdot (T\cdot d_{enc}^{2}+|\mathcal{M}|\cdot d_{enc}\cdot d_{K}+|\mathcal{M}| \cdot L\cdot n\cdot d_{s}^2))$.
\\
\textbf{Space complexity}.  The space of linear feature transformation in Eq.(\ref{eq:sdp_attn}) is $O(d_{enc}^{2})$. For attention in the patient representation and historical medication integration, the space is $O(d_{enc}\cdot d_{K})$. 
The GNN encoder (Eq.(\ref{eq:gin:agg})-(\ref{eq:gin:readout})) costs $O(n\cdot d_{s})$ space.
The matrix in Eq.(\ref{eq:crossmodal:predict}) costs $O(|\mathcal{M}|\cdot d_{K})$ space for the medication prediction task.
Assume $|\mathcal{M}|< d_{K}< d_{enc}$, the total space complexity is $O(d_{enc}^{2}+n\cdot d_{s})$.

\section{Experimental Study}
\label{sec:exp}
In this section, we describe the experimental setting~(\cref{sec:exp:setup}) and report the experiments in the following facets: 
(1) Compare the modeling effectiveness of \NLACMR with state-of-the-art drug recommendation approaches~(\cref{sec:exp:effect}).
(2) Conduct ablation studies to test the core components of \NLACMR~(\cref{sec:exp:ablation}).
(3) Test the effect of PLMs on \NLACMR~(\cref{sec:exp:various_PLMs}).
(4) Compare the impact of different numbers of visits on \NLACMR~ (\cref{sec:exp:various_visits}) and two strong baselines.
(5) Conduct a visualization study on the effectiveness of the alignment module proposed in our approach (\cref{sec:exp:visualization}). 
(6)  Investigate the sensitivity of \NLACMR regarding different hyper-parameter configurations
~(\cref{sec:exp:hyperparameter_sensitivity}).
(7) Compare the efficiency of \NLACMR with the deep-learning based baselines~(\cref{sec:exp:efficiency}).

\vspace{-0.2cm}
\subsection{Experimental Setup}
\label{sec:exp:setup}
\stitle{Baselines.} To comprehensively evaluate \NLACMR, we compare with 3 instance-based methods: \LR, \ECC~\cite{DBLP:conf/pkdd/ReadPHF09}, \LEAP~\cite{zhang2017leap}, and 7 longitudinal-based methods:  \RETAIN~\cite{DBLP:conf/nips/ChoiBSKSS16}, \GAMENet~\cite{luo2021ecnet}, \MICRON~\cite{DBLP:conf/ijcai/YangXGS21}, \SafeDrug~\cite{DBLP:conf/ijcai/YangXMGS21}, \COGNet~\cite{DBLP:conf/www/WuQJQW22}, \DrugRec~\cite{DBLP:conf/nips/Sun0LCW022} and  \MolRec~\cite{DBLP:conf/www/YangZWY23}.
\stitle{Datasets.} 
We use 3 public datasets: \mimicIII~\cite{johnson2016mimic}, \mimicIV~\cite{johnson2018mimic} and \eICU~\cite{pollard2018eicu}. 
For the \mimicIII and \mimicIV datasets, we follow the data processing of \DrugRec~\cite{DBLP:conf/nips/Sun0LCW022}, which collates the diagnosis, procedure, and medication records of patients. 
For the \eICU dataset, we only collect the diagnosis and medication records of the patients since the \eICU dataset itself does not include procedure records. 
Because the DDI between the medications in the \eICU dataset is not provided, we do not evaluate the DDI rate on \eICU. 
Table~\ref{exp:datasets} lists the profiles of the processed data.
For the MIMIC datasets, we choose the ATC Third Level code as the target label and extract its corresponding textual description from  DrugBank~\cite{DBLP:journals/nar/WishartFGLMGSJL18} as the textual medication description.
As the ATC code is not available in the \eICU dataset, we choose the generic therapeutic class (GTC) as the target label and obtain the textual description of the GTC code by engaging professionals to annotate its corresponding ATC Third Level code. The textual description of the ATC Third Level code is then adopted as the textual description of the GTC code.
We split train, validation, and test by 2/3, 1/6, and 1/6, respectively.

\begin{table}[h]
	\centering
        \small 
	\caption{Profile of datasets}	
	\vspace{-0.2cm}
	\label{exp:datasets}
			\begin{tabular}{l|r r r}
			\toprule
			{Items}  
   
   & \mimicIII   &\mimicIV  &\eICU
   \\ \midrule
   
	{\# of patients}		                   & 5,208 & 6,136 & 9,539
 \\

        {\# of visits }		      & 13,490 & 17,813  & 20,912
        \\ 
       {\# of diagnoses }  & 1,895 & 1,851 & 1,371
       \\
        {\# of procedures }	 & 1,378 & 4,001 & -
        \\
        {\# of symptoms }	 & 428 & 163  & - \\
                {\# of medications } &112   &121 &31 
                \\
        \midrule
        Avg {\# of visits }	 &2.59  &2.90 & 2.19
        \\
        Avg {\# of diagnoses } per visit 	 &10.24   &11.78 & 4.70\\
        Avg {\# of procedures } per visit &3.85  &2.18 & - \\
        Avg {\# of symptoms } per visit &7.67  &1.09  & - \\
        Avg {\# of medications } per visit &11.30   &6.68 & 5.98 \\

        \bottomrule
		
		\end{tabular}	
     \vspace{3pt}  
\end{table}

\vspace{-0.2cm}


\begin{table*}[ht]
	\centering
	\caption{Result comparison of different methods on three public datasets: \mimicIII, \mimicIV and \eICU}
	\vspace{-0.2cm}	
	\label{tab:exp:performance}
 
 \resizebox{\textwidth}{!}{

\renewcommand{\arraystretch}{1.05}
\begin{tabular}{l|c c c|c c c|c c c}
\toprule
   \multirow{2}{*}{Method}  & \multicolumn{3}{c|}{\mimicIII}   & \multicolumn{3}{c|}{\mimicIV} 
  & \multicolumn{3}{c}{\eICU}   
  \\  
  
   & \multicolumn{1}{c}{Jaccard $\uparrow$} & \multicolumn{1}{c}{F1 $\uparrow$} 
& {PRAUC $\uparrow$}  & \multicolumn{1}{c}{Jaccard $\uparrow$} 
& \multicolumn{1}{c}{F1 $\uparrow$} 
& {PRAUC $\uparrow$ } 
& \multicolumn{1}{c}{Jaccard $\uparrow$} 
& \multicolumn{1}{c}{F1 $\uparrow$} & {PRAUC $\uparrow$}

        \\ \midrule     

\LR  & $0.4896_{\pm 0.0025}$   &  $0.6491_{\pm 0.0024}$  & $0.7568_{\pm 0.0025}$  
&  $0.3844_{\pm 0.0028}$  & $0.5379_{\pm 0.0031}$  & $0.6568_{\pm 0.0036}$  
& $0.4199_{\pm 0.0027}$ & $0.5723_{\pm 0.0031}$ & $0.7515_{\pm 0.0032 }$
\\

\ECC    &$0.4799_{\pm 0.0022}$   & $0.6390_{\pm 0.0022}$  &$0.7572_{\pm 0.0026}$ 
 & $0.3680_{\pm 0.0041}$ & $0.5173_{\pm 0.0047}$ &  $0.6541_{\pm 0.0030}$ 
& $0.3786_{\pm 0.0036}$ & $0.5240_{\pm 0.0039}$ & $0.7498_{\pm 0.0036}$

\\

\LEAP                         
& $0.4465_{\pm 0.0037}$  &  $0.6097_{\pm 0.0036}$ & $0.6490_{\pm 0.0033}$  
& $0.3653_{\pm 0.0028}$ & $0.5201_{\pm 0.0033}$ & $0.5314_{\pm 0.0038}$ 
& $0.4514_{\pm 0.0035}$ & $0.5986 _{\pm 0.0036}$ & $0.6645_{\pm 0.0048}$
\\ \midrule 

\RETAIN                    
& $0.4780_{\pm 0.0036}$ & $0.6397_{\pm 0.0036}$ &$0.7601_{\pm 0.0035}$ 
& $0.3903_{\pm 0.0038}$  & $0.5471_{\pm 0.0040}$  & $0.6563_{\pm 0.0055}$ 
&  $0.4240_{\pm 0.0033}$ & $0.5723_{\pm 0.0034}$ & $0.7199 _{\pm 0.0036}$

\\

\GAMENet & $0.5039_{\pm 0.0021}$ & $0.6609_{\pm 0.0020}$ & $0.7632_{\pm 0.0027}$ 
& $0.3957_{\pm 0.0035}$ & $0.5525_{\pm 0.0041}$ & $0.6479_{\pm 0.0055}$ 
&  $\underline{0.4590_{\pm 0.0014}}$ & $\underline{0.6073_{\pm 0.0012}}$ & $0.7557_{\pm 0.0026}$ 

\\ 

\MICRON
& $0.5076_{\pm 0.0037}$ & $0.6634_{\pm 0.0035}$ & $0.7685_{\pm 0.0038}$ 
& $0.4009_{\pm 0.0044}$ & $0.5545_{\pm 0.0048}$ & $0.6584_{\pm 0.0043}$ 
&  $0.3741_{\pm 0.0042}$ & $0.5192_{\pm 0.0044}$ & $0.7551_{\pm 0.0039}$ 

\\ 

 \SafeDrug                     
& $0.5090_{\pm 0.0038}$ & $0.6664_{\pm 0.0033}$ & $0.7647_{\pm 0.0020}$
& $0.4082_{\pm 0.0026}$ & $0.5651_{\pm 0.0028}$ & $0.6495_{\pm 0.0036}$
& $0.4474_{\pm 0.0038}$ & $0.5964_{\pm 0.0032}$ & $0.7565_{\pm 0.0037}$

\\ 
 \COGNet                    
& $0.5134_{\pm 0.0027}$ & $0.6706_{\pm 0.0043}$ & $0.7677_{\pm 0.0013}$
& $0.4131_{\pm 0.0020}$ & $0.5660_{\pm 0.0019}$ & $0.6460_{\pm 0.0017}$
& $0.4588_{\pm 0.0020}$  & $0.5963_{\pm 0.0024}$ & $0.7263_{\pm 0.0020}$
\\

 \DrugRec                         
& $0.5220_{\pm 0.0034}$ & $0.6771_{\pm 0.0031}$ & $0.7720_{\pm 0.0036}$ 
& $0.4194_{\pm 0.0020}$ & $0.5713_{\pm 0.0022}$ & $0.6558_{\pm 0.0026}$
& $0.4508_{\pm 0.0030}$ & $0.5999_{\pm 0.0027}$ & $0.7496_{\pm 0.0030}$

\\ 

 \MolRec                
& $\underline{0.5278_{\pm 0.0033}}$ & $\underline{0.6825_{\pm 0.0030}}$ & $\underline{0.7736_{\pm 0.0036}}$
& $\underline{0.4197_{\pm 0.0030}}$ & $\underline{0.5772_{\pm 0.0032}}$ & $\underline{0.6644_{\pm 0.0046}}$
& $0.4476_{\pm 0.0031}$
& $0.5958_{\pm 0.0032}$
& $\underline{0.7575_{\pm 0.0033}}$


\\ 

 \midrule     
 \NLACMR       

& \cellcolor{LightCyan}\textbf{{$\textbf{0.5429}_{\pm 0.0027}$}}
& \cellcolor{LightCyan}\textbf{{$\textbf{0.6958}_{\pm 0.0024}$}}
& \cellcolor{LightCyan}\textbf{{$\textbf{0.7890}_{\pm 0.0016}$}}

& \cellcolor{LightCyan}\textbf{{$\textbf{0.4491}_{\pm 0.0044}$}}
& \cellcolor{LightCyan}\textbf{{$\textbf{0.6051}_{\pm 0.0042}$}}
& \cellcolor{LightCyan}\textbf{{$\textbf{0.6921}_{\pm 0.0045}$}} 

& \cellcolor{LightCyan}\textbf{{$\textbf{0.4787}_{\pm 0.0025}$}}
& \cellcolor{LightCyan}\textbf{{$\textbf{0.6262}_{\pm 0.0025}$}}
& \cellcolor{LightCyan}\textbf{{$\textbf{0.7650}_{\pm 0.0039}$}}

 \\ 

\bottomrule

\end{tabular}}

\end{table*}

\noindent\textbf{Evaluation Metrics}. We use the following metrics: 
(a) F1 score~(F1) - the harmonic mean of precision and recall; 
(b) Jaccard score~(Jaccard) - the ratio of the intersection to the union of predicted and true labels; 
(c) Precision Recall AUC~(PRAUC) - the area under the recall-precision curve, measuring the performance of a model across various recall levels. 

\noindent\textbf{Implementation and Reproducibility.}
Our experiments are performed using PyTorch 2.0.1. 
For drug molecule encoding, we adopt a 4-layer GIN~\cite{DBLP:conf/icml/SatorrasHW21} with a hidden embedding size of $d_s$ = 64. For each molecule graph, the 9-dimensional initial node features contain the atomic number and chirality, as well as other additional atom features, while the 3-dimensional edge features contain bond type, bond stereochemistry and whether the bond is conjugated. 
The size of text embeddings, denoted as $d_{enc}$, is 768. The projection layers are implemented as the MLP with 3 hidden layers of sizes [768, 256, 256]. We set the dropout ratio as 0.2.
The threshold value $\delta$ is set to 0.5. The loss weight $\alpha$ is set to 0.95.
We use Adam optimizer with a decaying learning rate to train the model for 110 epochs and the initial learning rate is set to 0.0005.
We use NVIDIA 3090 GPUs.
We consider five different PLMs and use public releases of \BioBERT~\cite{DBLP:journals/bioinformatics/LeeYKKKSK20}\footnote{\url{https://huggingface.co/dmis-lab/biobert-base-cased-v1.2}}, \SciBERT~\cite{DBLP:conf/emnlp/BeltagyLC19}\footnote{\url{https://huggingface.co/allenai/scibert_scivocab_uncased}}, \ClinicalBERT~\cite{lee2020clinical}\footnote{\url{https://huggingface.co/emilyalsentzer/Bio_ClinicalBERT}}, \PubBERT~\cite{DBLP:journals/health/GuTCLULNGP22}\footnote{\url{https://huggingface.co/microsoft/BiomedNLP-BiomedBERT-base-uncased-abstract-fulltext}}
and \BleuBERT~\cite{DBLP:conf/bionlp/PengYL19}\footnote{\url{https://huggingface.co/bionlp/bluebert_pubmed_mimic_uncased_L-12_H-768_A-12}}, and PLMs are frozen during the experiments.

\vspace{-0.2cm}
\subsection{Overall Performance}
\label{sec:exp:effect}
We compare \NLACMR with the baselines in terms of Jaccard, F1, and PRAUC on \mimicIII, \mimicIV and \eICU using \BioBERT~\cite{DBLP:journals/bioinformatics/LeeYKKKSK20} as the PLM. 
The comparison results in Table~\ref{tab:exp:performance} show that \NLACMR outperforms previous approaches regarding all the evaluation metrics, by a significant margin. 
The main reason for this is our effective integration of expert knowledge through the introduction of PLMs to extract patient and medication representations, followed by the alignment of the representations.
Additionally, our medication representations, which take into account both the chemical structures and textual descriptions, exhibit greater expressiveness compared to the representations employed in \MolRec and \DrugRec, which fail to leverage the textual medication descriptions for learning medication representations. 
The instance-based approaches such as \LR, \ECC, and \LEAP perform poorly and this is because these approaches fail to utilize longitudinal information in the EHR.
Among the longitudinal-based methods, \GAMENet incorporates additional graph information, while \SafeDrug and \MolRec utilize drug molecule structures and substructures, respectively. These enhancements contribute to further performance improvements.

\begin{figure}[t]
\hspace{-0.5cm}
\begin{center}
\includegraphics[width=0.90\columnwidth]{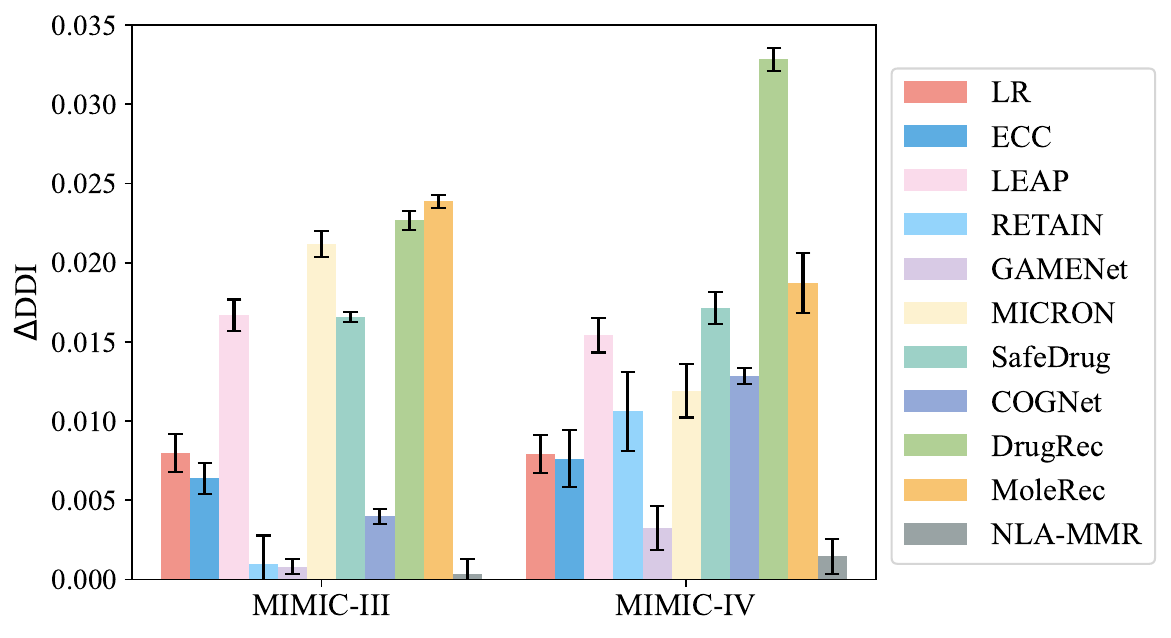}
\end{center}
\vspace*{-0.3cm}
\caption{ $\Delta{DDI}$ comparison of different methods on \mimicIII and \mimicIV 
}
\label{sec:exp:DDI}
\end{figure}

The average DDI rate of the ground truth medication combinations is 0.08238 and 0.07423 in \mimicIII and \mimicIV, respectively. We compute the DDI rate from the recommendation results of each method.  Then we report the difference between the DDI of each method and that of the ground truth (i.e., $\Delta DDI$) in Figure~\ref{sec:exp:DDI}. We can see that the DDI rate from the predicted results by \NLACMR has the smallest difference from that of the ground truth. This observation demonstrates that our proposed \NLACMR model proficiently simulates physicians' medication prescribing patterns.

\vspace{-0.3cm} 
\subsection{Ablation Study}
\label{sec:exp:ablation}
\begin{table}[t]
	\centering
        \small 
	\caption{Ablation study on \NLACMR}	
	\vspace{-0.2cm}
	\label{exp:tab:ablation_study}
			\begin{tabular}{l|c c c }
			\toprule
			\multirow{2}{*}{Model Variants}   & \multicolumn{3}{c}{\mimicIII}   \\  
			                      & {Jaccard} & {F1} & {PRAUC}  \\ \hline     
      		(a)	\textit{w/o} $E_{s}$             & 0.5368 & 0.6909   & 0.7872 \\
       	(b) \textit{w/o} $E_{f}$                   & 0.4747 & 0.6352 & 0.7394
        
       \\ \midrule
		(c) \textit{w/o} cross attention fusion           & 0.5348  & 0.6892   & 0.7815      \\
  
		(d) \textit{w/o} historical information        & 0.5303 &  0.6848   & 0.7794     \\ \midrule

  		 (e)  cross attention fusion \textit{w/o} $h_{d}$                      & 0.5413 & 0.6946 & 0.7867        \\

		 (f)  cross attention fusion \textit{w/o} $h_{p}$               & 0.5425 & 0.6955 & 0.7884
   
   \\
	 (g)  cross attention fusion \textit{w/o} $h_{s}$                  & 0.5415 & 0.6946 & 0.7883

  \\ \midrule
			  \textbf{Full version}           & \textbf{0.5429} & \textbf{0.6958} & \textbf{0.7890}     \\
     \bottomrule
		
		\end{tabular}	
  \vspace{0.2cm}
\end{table}

In this section, we conduct ablation studies to investigate the effectiveness of different components and settings of \NLACMR, including the drug representation module, cross-attention fusion, historical information integration mechanism, and using different types of textual descriptions in the EHRs for cross-attention fusion.
First, we ablate the structure-level representation and PLM-based functional representation in \NLACMR.
From Table~\ref{exp:tab:ablation_study}~(a)-(b), we can observe that conducting the recommendation solely based on either the structural or textual medication representation results in performance degradation, which confirms the importance of incorporating both structural information and textual knowledge in the medication modality.

Second, to exploit the effectiveness of the cross-attention fusion in the patient representation module and historical information integration technique, we remove these two components, respectively. As shown in Table~\ref{exp:tab:ablation_study}~(c)-(d), only leveraging the feature in Eq.~\eqref{eq:dps_feature} as the patient representation will degrade the performance of our proposed model to a large extent, showing our model can benefit from fusing different types of texts in the EHRs. 
Besides, we notice that disabling the historical information hurts performance significantly on all evaluation metrics.

Third, we investigate the importance of different types of text from the patient modality used in the cross-attention fusion module in Table~\ref{exp:tab:ablation_study}~(e)-(g). 
For each ablation, we ignore one type of textual description for conducting cross-attention fusion in Eq.~\eqref{eq:sdp_attn}. 
We observe that implementing the cross-attention with two types of textual description is also beneficial and the best performance is achieved with all the types of textual description. 

\vspace{-0.2cm}
\subsection{Effect of Pretrained Language Models}
\label{sec:exp:various_PLMs}
\begin{table*}[ht]
	\centering
	\caption{The influence of using various PLMs on \NLACMR}
	\vspace{-0.2cm}	
	\label{tab:exp:PLM_performance}
\resizebox{\textwidth}{!}{
\renewcommand{\arraystretch}{1.05}
\begin{tabular}{l|c c c|c c c|c c c}
\toprule
   \multirow{2}{*}{PLM}  & \multicolumn{3}{c|}{\mimicIII}   & \multicolumn{3}{c|}{\mimicIV}  
   & \multicolumn{3}{c}{\eICU}  \\  
   & {Jaccard $\uparrow$} & {F1 $\uparrow$} 
& {PRAUC $\uparrow$}  & {Jaccard $\uparrow$} 
& {F1 $\uparrow$} 
& {PRAUC $\uparrow$}  
& {Jaccard $\uparrow$} & {F1 $\uparrow$} 
& {PRAUC $\uparrow$} 
       \\ \midrule

  \BioBERT

& \cellcolor{LightCyan}\textbf{{$\textbf{0.5429}_{\pm 0.0027}$}}
& \cellcolor{LightCyan}\textbf{{$\textbf{0.6958}_{\pm 0.0024}$}}
& \cellcolor{LightCyan}\textbf{{$\textbf{0.7890}_{\pm 0.0016}$}}

& \cellcolor{LightCyan}\textbf{{$\textbf{0.4491}_{\pm 0.0044}$}}
& \cellcolor{LightCyan}\textbf{{$\textbf{0.6051}_{\pm 0.0042}$}}
& \cellcolor{LightCyan}\textbf{{$\textbf{0.6921}_{\pm 0.0045}$}} 

& \cellcolor{LightCyan}\textbf{{$\textbf{0.4787}_{\pm 0.0025}$}}
& \cellcolor{LightCyan}\textbf{{$\textbf{0.6262}_{\pm 0.0025}$}}

& ${0.7650}_{\pm 0.0039}$
       
\\ 
\PubBERT       
& $0.5375_{\pm 0.0024}$
& $0.6909_{\pm 0.0020}$
& $0.7849_{\pm 0.0017}$
& $0.4471_{\pm 0.0044}$
& $0.6028_{\pm 0.0043}$
& $0.6891_{\pm 0.0043}$
& $0.4762_{\pm 0.0023}$ 
& $0.6240_{\pm 0.0022}$
& \cellcolor{LightCyan}\textbf{{$\textbf{0.7662}_{\pm 0.0042}$}}

\\
  \ClinicalBERT       
& $0.5382_{\pm 0.0025}$
& $0.6917_{\pm 0.0021}$
& $0.7838_{\pm 0.0016}$ 
& $0.4483_{\pm 0.0044}$
& $0.6037_{\pm 0.0042}$
& $0.6880_{\pm 0.0043}$
& $0.4755_{\pm 0.0018}$
& $0.6227_{\pm 0.0017}$
& $0.7649_{\pm 0.0038}$
\\ 
  \BleuBERT     
& $0.5296_{\pm 0.0026}$
& $0.6842_{\pm 0.0022}$
& $0.7772_{\pm 0.0016}$
& $0.4377_{\pm 0.0042}$
& $0.5946_{\pm 0.0042}$
& $0.6785_{\pm 0.0042}$
& $0.4712_{\pm 0.0026}$
& $0.6186_{\pm 0.0026}$
& $0.7606_{\pm 0.0042}$
\\ 
  \SciBERT            
& $0.5336_{\pm 0.0032}$
& $0.6878_{\pm 0.0027}$
& $0.7779_{\pm 0.0019}$
& $0.4399_{\pm 0.0053}$
& $0.5961_{\pm 0.0050}$
&  $0.6810_{\pm 0.0050}$
& $0.4719_{\pm 0.0023}$ 
& $0.6195_{\pm 0.0022}$
& $0.7632_{\pm 0.0030}$
\\
\bottomrule
\end{tabular}}
\end{table*}
To assess the impact of different PLMs on \NLACMR, we use five public releases of PLMs, including \BioBERT~\cite{DBLP:journals/bioinformatics/LeeYKKKSK20}, \SciBERT~\cite{DBLP:conf/emnlp/BeltagyLC19}, \ClinicalBERT~\cite{lee2020clinical}, \PubBERT~\cite{DBLP:journals/health/GuTCLULNGP22} and \BleuBERT~\cite{DBLP:conf/bionlp/PengYL19}. 
Among these PLMs, \BioBERT is initialized with the standard BERT model and then continues pretraining using PubMed abstracts, while \PubBERT is pretrained from scratch using these texts.
\ClinicalBERT conducts continual pretraining from \BioBERT with clinical notes from \mimicIII~\cite{johnson2016mimic}.
\BleuBERT mixes PubMed and \mimicIII~\cite{johnson2016mimic} to conduct continual pretraining from BERT.
\SciBERT is pretrained from scratch using biomedicine and computer science articles.

Table~\ref{tab:exp:PLM_performance} compares the performance of \NLACMR using five different PLMs. 
The evaluation measures for the \NLACMR implemented with all five PLMs are much higher than those for the best baseline \MolRec, which confirms the general applicability and superiority of our proposed approach. 
We observe that the model achieves the best performance when using \BioBERT as the PLM.
In particular, even though clinical notes in \mimicIII are relevant to the domain of diagnosis and procedure texts in our drug recommendation task, adding them to the pretraining corpus does not bring any advantage, as is evident by the results of \ClinicalBERT and \BleuBERT.
This reveals that mixed-domain pre-training of the texts related to the patients and drugs is not beneficial to the CMR task, 
which supports the assumption that textual descriptions of the patient and drugs are different from the two modalities.
However, the PLM such as \SciBERT mixing out-of-domain texts from science articles during the pretraining procedure leads to worse performance. 
\vspace{-0.2cm}
\subsection{Effect of Historical Visits}
\label{sec:exp:various_visits}

\begin{figure}[t]
    \centering
    \includegraphics[width=0.50\columnwidth]{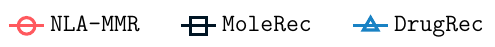}
\vspace{-0.2cm}

\begin{center}
\begin{tabular}[t]{c}
    
\hspace{-0.6cm}
   \subfigure[Jaccard]{
     \includegraphics[width=0.34\columnwidth]{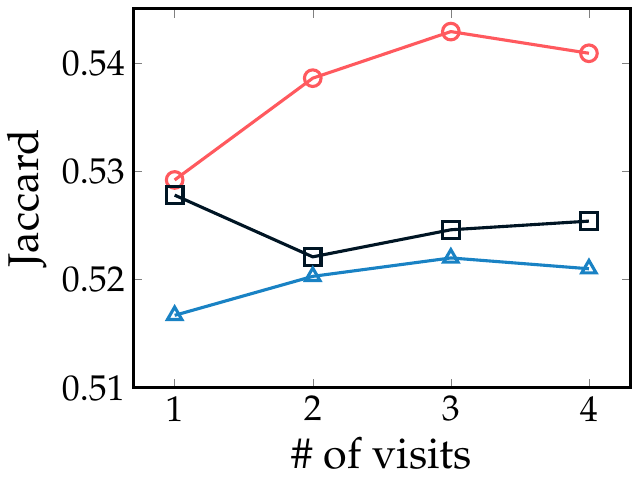} 
       \label{fig:visits:jaccard}
    }
    \hspace{-0.3cm}
    \subfigure[F1]{
      \includegraphics[width=0.34\columnwidth]{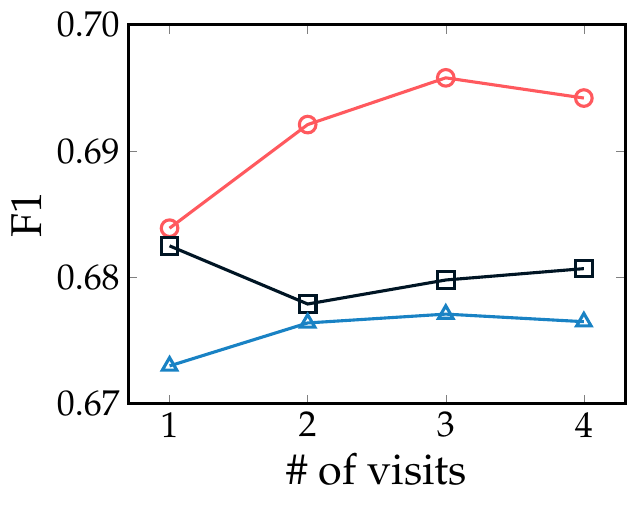} 
        \label{fig:visits:fone}
    }
    \hspace{-0.3cm}
    \subfigure[PRAUC]{
      \includegraphics[width=0.34\columnwidth]{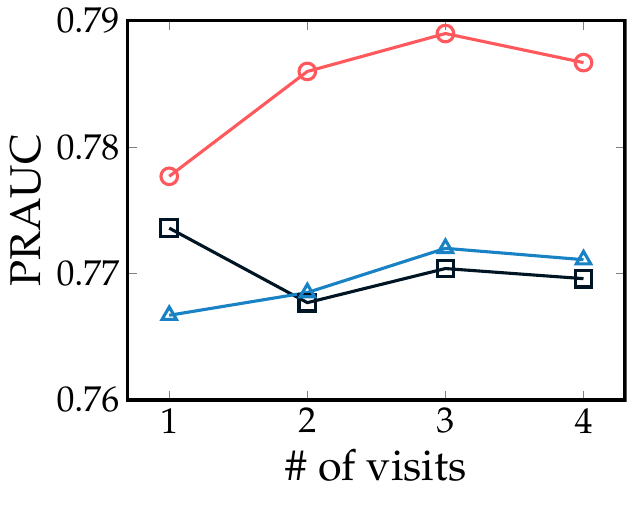} 
        \label{fig:visits:PRAUC}
    }
\end{tabular}
\end{center}
\vspace*{-0.3cm}
\caption{The impact of \# of visits on \mimicIII for two strong baselines and \NLACMR using \BioBERT as PLM}
\label{fig:exp:visit}
\vspace*{-0.05cm}
\end{figure}

To evaluate the impact of the number of historical visits on the model performance, we test our approach with different numbers of visits on \mimicIII using \BioBERT. 
As a comparison, we include the two strongest baselines \MolRec and \DrugRec, which also incorporate historical information.
Figure~\ref{fig:exp:visit} shows the Jaccard, F1, and PRAUC of \NLACMR and the two baselines when setting the number of visits in $\{1, 2, 3, 4\}$.
We observe that \NLACMR achieves relatively better performance with more visits, while the performance of \MolRec almost stays flat and \DrugRec consistently exhibits poor performance. 
This indicates that the design of the historical information integration mechanism can effectively incorporate prescription information of historical visits into the CMR task, thereby directly enhancing the accuracy of drug predictions.
On the other hand, \MolRec relies on an RNN-based mechanism, which may not be as adept at capturing and utilizing the relevant information from past visits, leading to its relatively stagnant performance.
Note that both \NLACMR and \DrugRec achieve the best performance by 
utilizing the historical information in the last two visits. 
It suggests that outdated information may not provide useful guidance and could potentially lead to misleading prescriptions in current drug predictions.

\vspace{-0.3cm}
\subsection{{Visualization}}
\label{sec:exp:visualization}
\begin{figure}[t]
\begin{center}
\begin{tabular}[t]{c}
\hspace{-0.5cm}
\centering
   \subfigure[MIMIC-III]{
     \includegraphics[width=0.47\columnwidth]{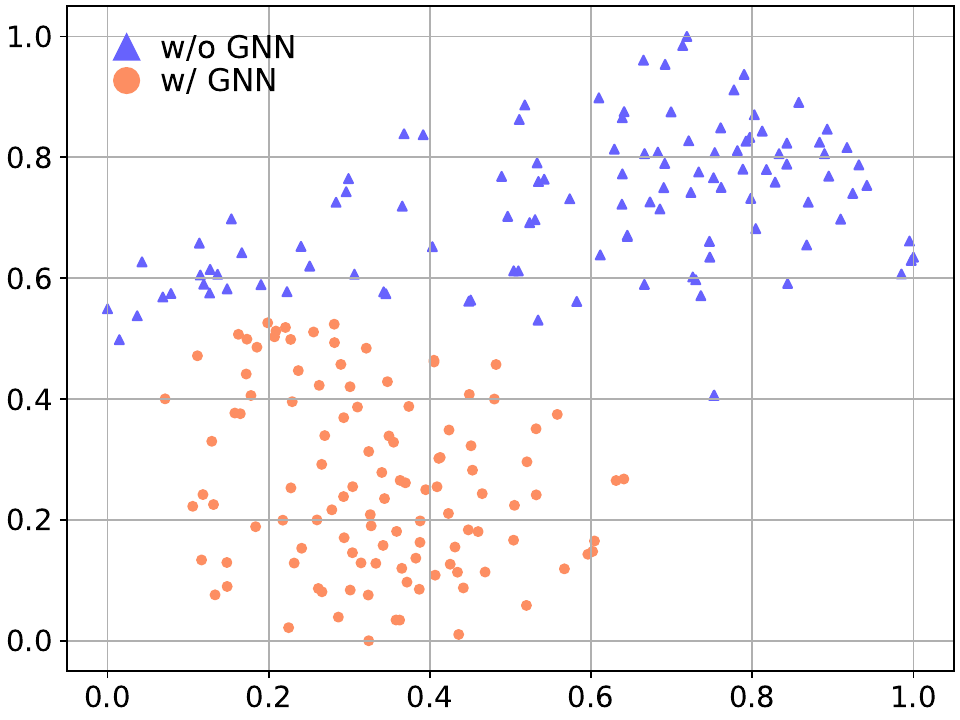} 
       \label{fig:visualize_med_3}
    }
       \hspace{-0.3cm}
    \subfigure[MIMIC-IV]{
      \includegraphics[width=0.47\columnwidth]{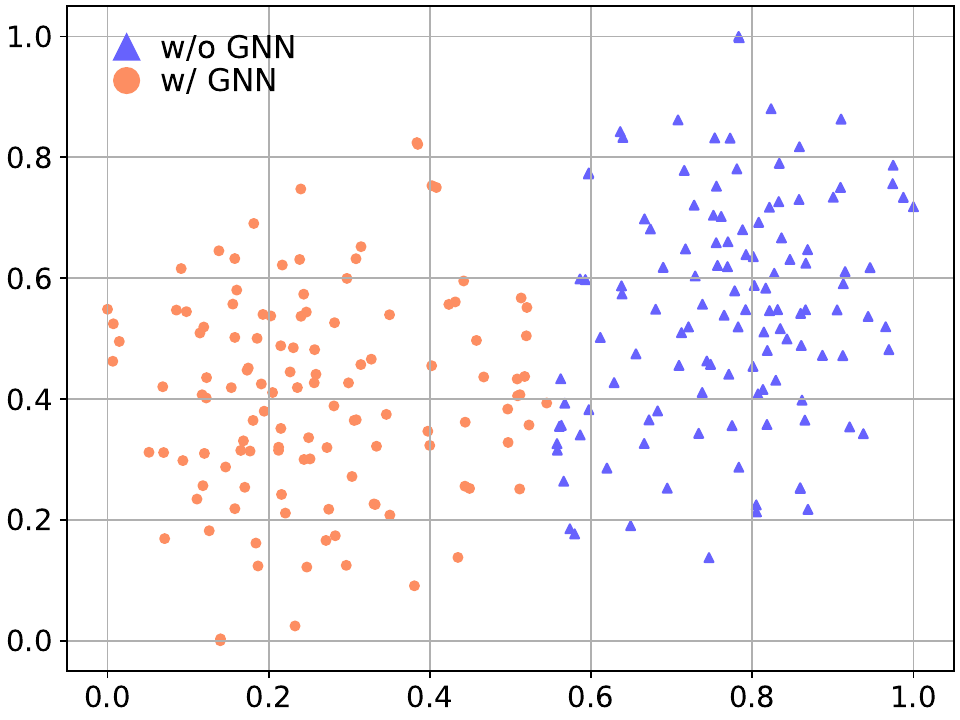} 
       \label{fig:visualize_med_4}
    }
    
\end{tabular}
\end{center}
\vspace*{-0.4cm}
\caption{Visualization of GNN-enhanced and text-only based medication representations using t-SNE
}
\label{fig:casestudy:vis_med}
\vspace*{-0.2cm}
\end{figure}

Recall that in the medication representation module, we incorporate the PLM-based functional representation with GNN-based structure-level representation as the medication representation (Eq.~\eqref{eq:concat_m}).
In this section, we conduct a case study to explore whether the structure-level representation can capture distinct information beyond the textual medication descriptions, and thus providing extra assistance for the CMR task.
First, we compare the two drug representations, the pure textural representation $\bm{E}_f$, and the representation $\bm{E}_m$ incorporated extra structural feature by GNN.    
Figure~\ref{fig:casestudy:vis_med} shows the visualization of the two representations, i.e., w/o GNN and w/ GNN, on the \mimicIII and \mimicIV by t-SNE~\cite{van2008visualizing}, where \BioBERT is used as the PLM.
We observe that the distributions of the representations w/o and w/ GNN are different. 
It is noticed that the \NLACMR model can learn the chemical properties of drugs because the 
incorporation of structural information. 
The drug representations derived from chemical structures and textual descriptions exhibit noticeable distinctions when compared to text-based drug representations in the embedding space.

\vspace*{-0.1cm}
\begin{figure}[t]
\begin{center}
\begin{tabular}[t]{c}
\hspace{-0.5cm}
   \subfigure[Before Training]{
     \label{fig:casestudy:vis_align:before}
     \includegraphics[width=0.48\columnwidth]{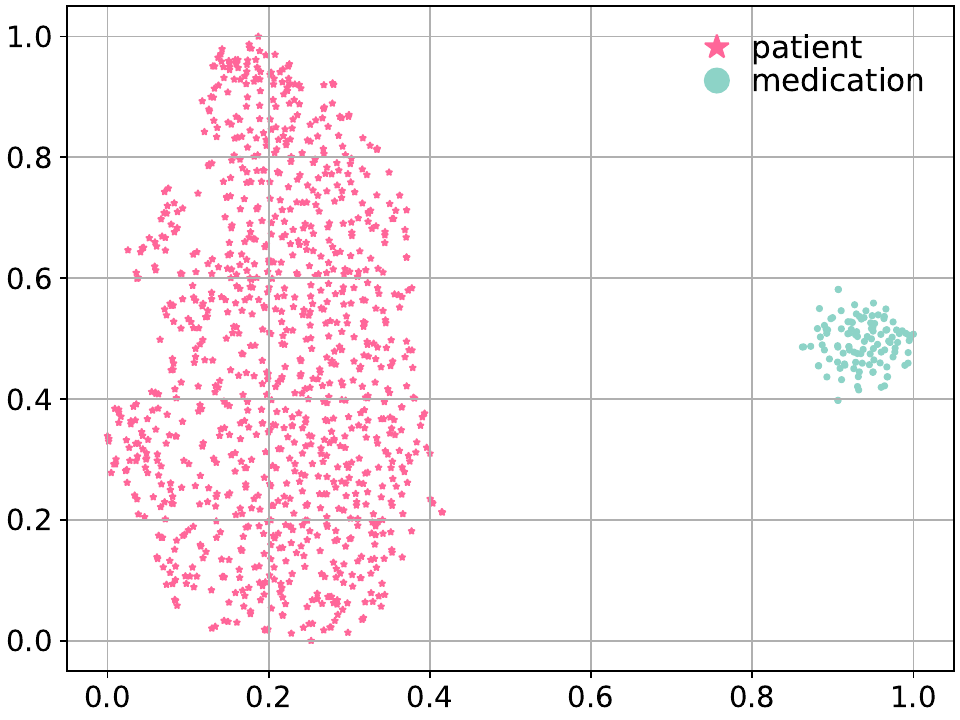} 
    }
       \hspace{-0.3cm}
    \subfigure[After Training]{
      \label{fig:casestudy:vis_align:after}
      \includegraphics[width=0.48\columnwidth]{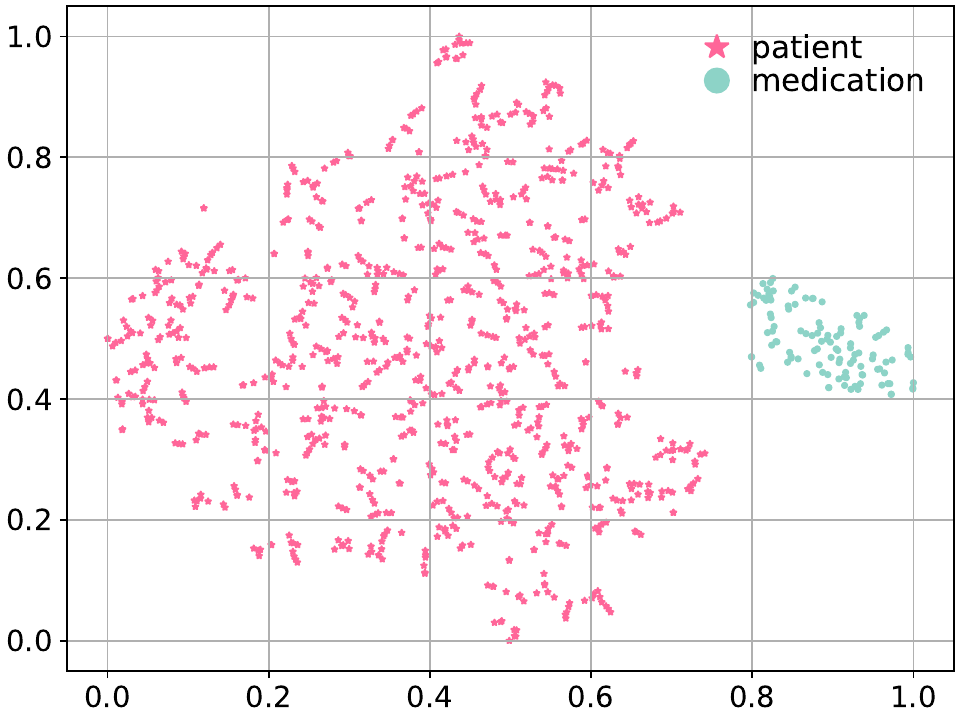} 
    }
    
\end{tabular}
\end{center}
\vspace{-10pt}
\caption{Visualization of alignment between the representations from the patients and medications}
\label{fig:casestudy:vis_align}
\vspace*{-0.1cm}
\end{figure}

Furthermore, to gain a better understanding of the multi-modal alignment module in \NLACMR, we visualize the representation of 1000 patient samples and the representation of all the 112 drugs, generated by \NLACMR before and after training.
Figure~\ref{fig:casestudy:vis_align} presents the two representations of \mimicIII by t-SNE, where \BioBERT is used as the PLM. 
We observe that before training the representations of patient and drug are two separate manifolds (Figure~\ref{fig:casestudy:vis_align:before}), while after the training, the two manifolds are closer in the latent space and present a trend of integration (Figure~\ref{fig:casestudy:vis_align:after}).
This visualization confirms that our multi-modal alignment module is capable of aligning the representations of patients and drugs.

\vspace{-0.2cm}
\subsection{Hyper-parameter Sensitivity}
\label{sec:exp:hyperparameter_sensitivity}
In this section, we test the parameter sensitivity of \NLACMR on the \mimicIII and \mimicIV datasets using \BioBERT as the PLM.
\\
\textbf{Impact of $batch$.} 
We conduct experiments with $lr$ = 5e-4 and $\alpha$ = 0.95. 
In Figure\ref{fig:exp:hyper} (a)-(c), we vary $batch$ from 2 to 64 and show the performance of \NLACMR. Overall, the performance of \NLACMR remains stable across different batch sizes. 
The performance gap of \NLACMR with varied batch size is 0.0123 and 0.0104 for \mimicIII and \mimicIV, respectively.  
Notably, the optimal batch size for \mimicIII and \mimicIV are 8 and 4, respectively.
\\
\textbf{Impact of $lr$.}
We conduct experiments with $\alpha$ = 0.95 and $batch$ = 8. In Figure \ref{fig:exp:hyper} (d)-(f), we vary $lr$ in \{1e-5, 5e-5, 1e-4, 5e-4, 1e-3, 5e-3, 1e-2\} and show the performance of \NLACMR. The results indicate that our model is sensitive to the learning rate and the optimal learning rate is 5e-4 across two datasets.
\\
\textbf{Impact of $\alpha$.}
We conduct experiments with $lr$ = 5e-4 and $batch$ = 8. In Figure \ref{fig:exp:hyper} (g)-(i), we set $\alpha$=0 and the significant performance degradation confirms the greater importance of the bce loss.
Another observation is that our model is stable with varied $\alpha$ from 0.90 to 1.0 and the optimal value for $\alpha$ is 0.95 across two datasets.

\vspace{-0.2cm}
\subsection{Efficiency}
\label{sec:exp:efficiency}
We compare the training and inference time of \NLACMR with 8 deep learning-based baselines for \mimicIII and \mimicIV datasets on a Quadro RTX 800 GPU in Table \ref{tab:efficiency}.
The results demonstrate that our \NLACMR consistently outperforms most of the baselines in terms of training time across two datasets.
Additionally, its inference time is comparable to that of \RETAIN, \SafeDrug, and \GAMENet, and better than the other baselines across two datasets. 
Therefore, we conclude that \NLACMR is a viable solution for real deployment.

\begin{figure}[t]
    \centering
    \includegraphics[width=0.50\columnwidth]{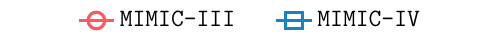}
\vspace{-0.4cm}

\begin{center}
\begin{tabular}[t]{c}
    
\hspace{-0.6cm}
   \subfigure[Jaccard]{
     \includegraphics[width=0.33\columnwidth]{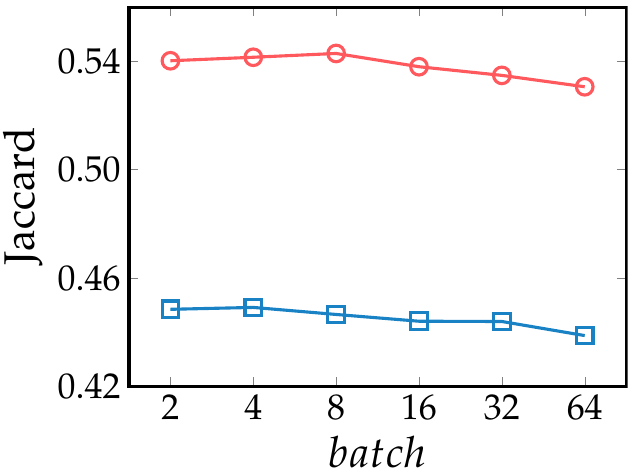} 
       \label{fig:batch:ja}
    }
    
    \hspace{-0.3cm}
    \subfigure[F1]{
      \includegraphics[width=0.33\columnwidth]{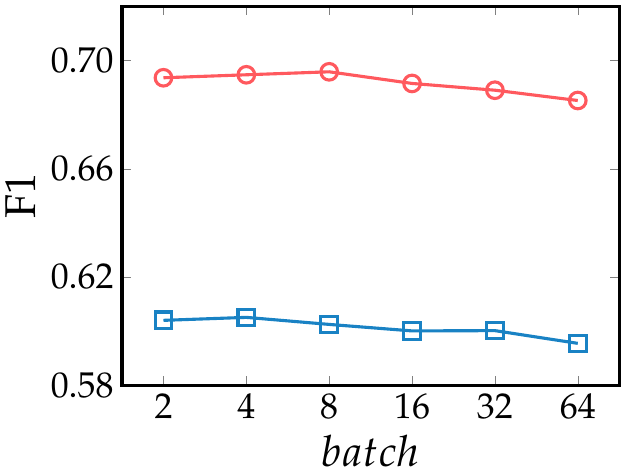} 
        \label{fig:batch:f1}
    }
    \hspace{-0.3cm}
    \subfigure[PRAUC]{
      \includegraphics[width=0.33\columnwidth]{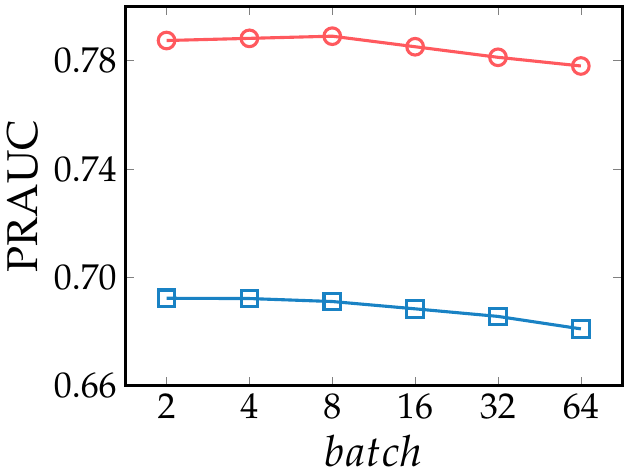} 
          \label{fig:batch:prauc}
    }
\end{tabular}
\end{center}

\label{fig:exp:visits}
\vspace*{-0.75cm}
\end{figure}

\begin{figure}[t]
    \centering

\begin{center}
\begin{tabular}[t]{c}
    
\hspace{-0.60cm}
   \subfigure[Jaccard]{
     \includegraphics[width=0.33\columnwidth]{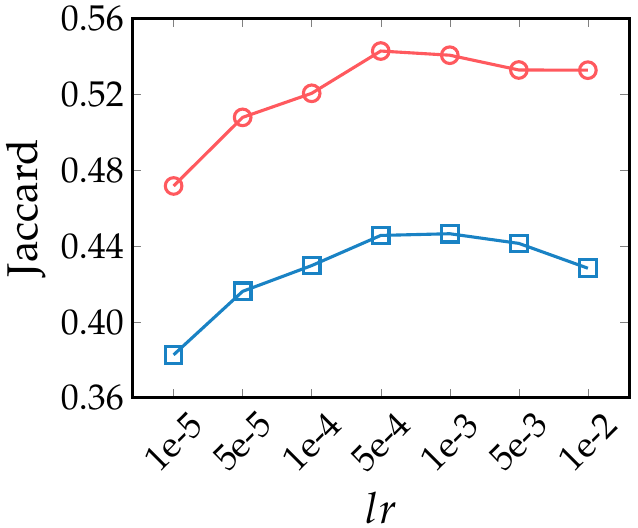} 
        \label{fig:lr:ja}
    }
    
    \hspace{-0.3cm}
    \subfigure[F1]{
      \includegraphics[width=0.33\columnwidth]{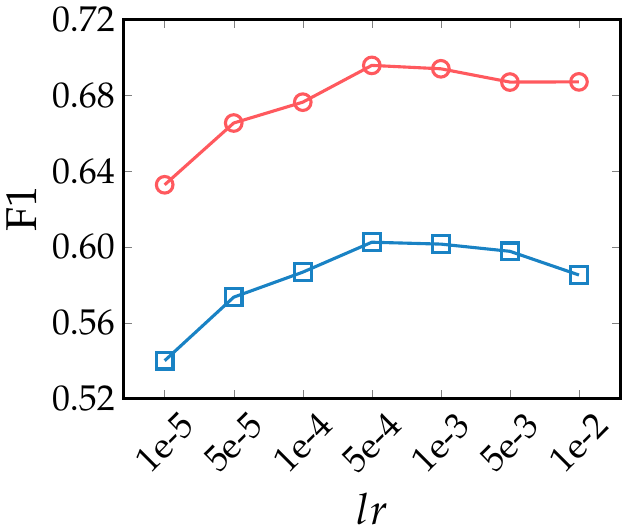} 
   \label{fig:lr:f1}
    }
    \hspace{-0.3cm}
    \subfigure[PRAUC]{
      \includegraphics[width=0.33\columnwidth]{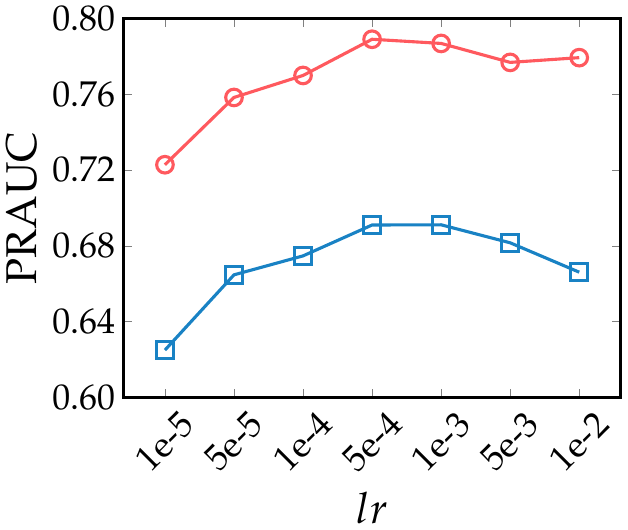} 
   \label{fig:lr:prauc}
    }
\end{tabular}
\end{center}
\label{fig:exp:lr}
\vspace*{-0.75cm}
\end{figure}

\begin{figure}[t]
    \centering

\begin{center}
\begin{tabular}[t]{c}
    
\hspace{-0.6cm}
   \subfigure[Jaccard]{
     \includegraphics[width=0.33\columnwidth]{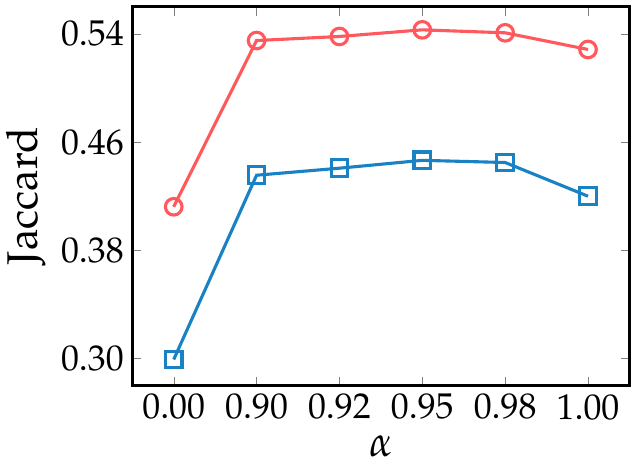} 
       \label{fig:weight:ja}
    }
    
    \hspace{-0.3cm}
    \subfigure[F1]{
      \includegraphics[width=0.33\columnwidth]{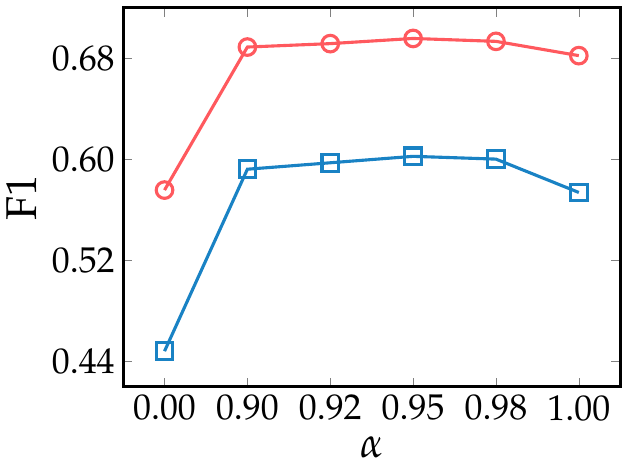} 
        \label{fig:weight:f1}
    }
    \hspace{-0.3cm}
    \subfigure[PRAUC]{
      \includegraphics[width=0.33\columnwidth]{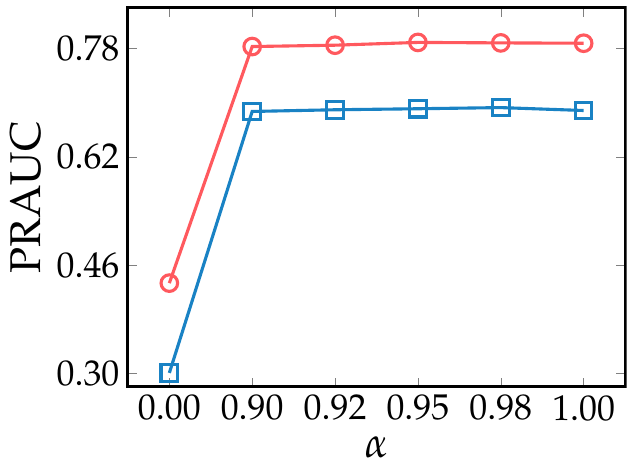} 
        \label{fig:weight:prauc}
    }
\end{tabular}
\end{center}
\vspace*{-0.4cm}
\caption{The impact of $batch$, $lr$, and $\alpha$ on \mimicIII and \mimicIV using \BioBERT as PLM}
\label{fig:exp:hyper}
\vspace*{-0.1cm}
\end{figure}

\begin{table}[htbp]
\small 
\caption{Efficiency comparison on \mimicIII and \mimicIV}
\label{tab:efficiency}
\begin{tabular}{l|ll|ll}
\toprule
Model &\multicolumn{2}{c|}{Training time~(s/Epoch)} &\multicolumn{2}{c}{Testing time~(s)}  \\ 
&\mimicIII &\mimicIV &\mimicIII &\mimicIV \\ 
\midrule
\LEAP & 317.24$_{\pm 9.27}$ & 422.83$_{\pm 14.70}$  & 26.55$_{\pm 0.44}$ & 28.26$_{\pm 0.16}$ \\
\RETAIN   & 45.81$_{\pm 2.82}$ & 54.44$_{\pm 2.10}$  & 3.78$_{\pm 0.10}$ & 5.62$_{\pm 0.27}$ \\
\GAMENet & 140.11$_{\pm 9.23}$ & 185.05$_{\pm 0.89}$  & 7.50$_{\pm 0.37}$ &10.69$_{\pm 0.04}$ \\
\MICRON  & 87.63$_{\pm 1.61}$ & 105.24$_{\pm 1.14}$  & 11.55$_{\pm 0.49}$ & 14.93$_{\pm 0.03}$  \\
\SafeDrug  & 194.47$_{\pm 21.19}$ & 303.54$_{\pm 13.03}$ & 7.64$_{\pm 1.06}$ & 9.76$_{\pm 0.47}$ \\
\COGNet  & 123.77$_{\pm 1.86}$ & 135.35$_{\pm 1.70}$ & 92.22$_{\pm 1.57}$ & 105.80$_{\pm 0.22}$  \\
\DrugRec  & 461.86$_{\pm 9.37}$ & 605.77$_{\pm 16.15}$ & 26.78$_{\pm 0.17}$ & 33.93$_{\pm 1.76}$  \\
\MolRec   & 479.22$_{\pm 10.89}$ & 551.38$_{\pm 2.78}$ & 27.54$_{\pm 1.95}$ &  28.81$_{\pm 0.07}$ \\
\NLACMR  & 50.83$_{\pm 3.25}$ & 62.98$_{\pm 2.34}$  & 7.67$_{\pm 0.21}$ &  10.50$_{\pm 0.28}$ \\

\bottomrule
\end{tabular}
\end{table}

\section{Related Work}
\label{sec:rw}
\subsection{Molecular Representation Learning}


Molecular representation learning aims to utilize deep learning models to encode the input molecules as numerical vectors, which preserve the structural and property information about the molecules and serve as feature vectors for downstream chemical applications. 
Existing medication recommendation approaches~\cite{DBLP:conf/nips/Sun0LCW022,DBLP:conf/ijcai/YangXMGS21,DBLP:conf/www/YangZWY23} mainly model the chemical structure of molecules through one-dimensional description~\cite{DBLP:journals/mlst/KrennHNFA20,huang2020caster}, two-dimensional molecular graphs~\cite{DBLP:conf/nips/DuvenaudMABHAA15,DBLP:conf/nips/LiuDL19,NEURIPS2020_94aef384,10.1093/bioinformatics/btac039,li2024neural} and three-dimensional geometric graphs~\cite{jiao2023energy,huang2022equivariant,NEURIPS2022_3bdeb28a}.
Besides the chemical structural information, we have textual medication descriptions that are available on PubChem~\cite{DBLP:journals/nar/00020CGHHLSTYZ021} and DrugBank~\cite{DBLP:journals/nar/WishartFGLMGSJL18}. 
It is observed that these textual medication descriptions offer a comprehensive perspective on the functional aspects of molecules and explain their therapeutic applications.

In this paper, we aim to incorporate textual knowledge into molecular representation learning.


\comment{
\begin{itemize}
\color{red}{
       \item rewrite intro (especially the first sentence in each paragraph) E1
    \item method (rewrite) E3
    \item do visualization (dsp/m =>Original PLM-based embedding => clip-> adapt to the CMR task
    ; GNN-integration)  (simple coding: validation/) C
    \item the framework figure 
    (1.Left to right pipeline 2.clear input texts => embedding procedure 3.test case  4.PLM logo)
        }
    
\end{itemize}

}

\vspace{-0.25cm}
\subsection{Medication Recommendation}
Existing approaches can be broadly classified into two main categories: instance-based methods and longitudinal methods.
Instance-based methods~\cite{zhang2017leap,DBLP:journals/bdr/GongWWWL21} only take information of the current visit as input. 
For example, \LEAP~\cite{zhang2017leap} formulates the medication recommendation task into a sequential decision-making process and adopts a multi-instance multi-label learning framework to generate the medication recommendations based on the patient’s current diagnosis information.

In contrast, longitudinal methods~\cite{DBLP:conf/kdd/Le0V18, DBLP:conf/cikm/WangRCR0R19,DBLP:conf/ijcai/YangXGS21,DBLP:conf/aaai/ShangXMLS19, DBLP:conf/kdd/WangZHZ18, DBLP:conf/ijcai/YangXMGS21} use the historical information of patients and explore sequential dependency within clinical visits.
Most of them model longitudinal patient information using Recurrent Neural Networks.
\GAMENet~\cite{DBLP:conf/aaai/ShangXMLS19} and \SafeDrug~\cite{DBLP:conf/ijcai/YangXMGS21} further incorporated the BioKG and DDI, respectively, to improve model performance. 
Despite significant progress made in using structured domain knowledge in the CMR task, these methods suffer from the inherent biases and information loss in the structured data caused by the preprocessing stage.
This paper aims to encode the raw texts in patient and drug modalities by using powerful PLMs as the foundation block to generate their representations, which is an effective way of integrating multimodal information in multimodal recommendation systems~\cite{mu2022learning,DBLP:journals/corr/abs-2302-04473}.
\vspace{-0.2cm}
\section{Conclusion}
\label{sec:conclusion}
In this paper, we consider the patient and medication as two distinct modalities and design a multi-modal framework \NLACMR to learn their representations according to the alignment. Besides, we investigate the potential of PLMs to extract domain knowledge in the textual descriptions from the patient and medication modalities. In the medication modality, we incorporate both the chemical structures and medication descriptions to enhance molecular representation learning.
Extensive experiments on three public datasets show the effectiveness and efficiency of our proposed \NLACMR.

\vspace{-0.2cm}
\section*{ACKNOWLEDGEMENT}
Yu Rong and Kangfei Zhao are the corresponding authors. This research was partially supported by the CUHK Stanley Ho Big Data Decision Analytics Research Centre. Kangfei Zhao is supported by National Key Research and Development Plan, No. 2023YFF0725101.
\bibliographystyle{ACM-Reference-Format}
\bibliography{sample-base}


\begin{thebibliography}{41}


\ifx \showCODEN    \undefined \def \showCODEN     #1{\unskip}     \fi
\ifx \showDOI      \undefined \def \showDOI       #1{#1}\fi
\ifx \showISBNx    \undefined \def \showISBNx     #1{\unskip}     \fi
\ifx \showISBNxiii \undefined \def \showISBNxiii  #1{\unskip}     \fi
\ifx \showISSN     \undefined \def \showISSN      #1{\unskip}     \fi
\ifx \showLCCN     \undefined \def \showLCCN      #1{\unskip}     \fi
\ifx \shownote     \undefined \def \shownote      #1{#1}          \fi
\ifx \showarticletitle \undefined \def \showarticletitle #1{#1}   \fi
\ifx \showURL      \undefined \def \showURL       {\relax}        \fi
\providecommand\bibfield[2]{#2}
\providecommand\bibinfo[2]{#2}
\providecommand\natexlab[1]{#1}
\providecommand\showeprint[2][]{arXiv:#2}

\bibitem[Beltagy et~al\mbox{.}(2019)]%
        {DBLP:conf/emnlp/BeltagyLC19}
\bibfield{author}{\bibinfo{person}{Iz Beltagy}, \bibinfo{person}{Kyle Lo}, {and} \bibinfo{person}{Arman Cohan}.} \bibinfo{year}{2019}\natexlab{}.
\newblock \showarticletitle{SciBERT: {A} Pretrained Language Model for Scientific Text}. In \bibinfo{booktitle}{\emph{In Proc. {EMNLP-IJCNLP}}}. \bibinfo{publisher}{ACL}, \bibinfo{pages}{3613--3618}.
\newblock


\bibitem[Bian et~al\mbox{.}(2023)]%
        {DBLP:conf/icde/BianJLXRSKM023}
\bibfield{author}{\bibinfo{person}{Tian Bian}, \bibinfo{person}{Yuli Jiang}, \bibinfo{person}{Jia Li}, \bibinfo{person}{Tingyang Xu}, \bibinfo{person}{Yu Rong}, \bibinfo{person}{Yi Su}, \bibinfo{person}{Timothy C.~Y. Kwok}, \bibinfo{person}{Helen Meng}, {and} \bibinfo{person}{Hong Cheng}.} \bibinfo{year}{2023}\natexlab{}.
\newblock \showarticletitle{Decision Support System for Chronic Diseases Based on Drug-Drug Interactions}. In \bibinfo{booktitle}{\emph{International Conference on Data Engineering}}. \bibinfo{publisher}{{IEEE}}, \bibinfo{pages}{3467--3480}.
\newblock


\bibitem[Choi et~al\mbox{.}(2016)]%
        {DBLP:conf/nips/ChoiBSKSS16}
\bibfield{author}{\bibinfo{person}{Edward Choi}, \bibinfo{person}{Mohammad~Taha Bahadori}, \bibinfo{person}{Jimeng Sun}, \bibinfo{person}{Joshua Kulas}, \bibinfo{person}{Andy Schuetz}, {and} \bibinfo{person}{Walter~F. Stewart}.} \bibinfo{year}{2016}\natexlab{}.
\newblock \showarticletitle{{RETAIN:} An Interpretable Predictive Model for Healthcare using Reverse Time Attention Mechanism}. In \bibinfo{booktitle}{\emph{Proc. {NeurIPS}}}. \bibinfo{pages}{3504--3512}.
\newblock


\bibitem[Duvenaud et~al\mbox{.}(2015)]%
        {DBLP:conf/nips/DuvenaudMABHAA15}
\bibfield{author}{\bibinfo{person}{David Duvenaud}, \bibinfo{person}{Dougal Maclaurin}, \bibinfo{person}{Jorge Aguilera{-}Iparraguirre}, \bibinfo{person}{Rafael G{\'{o}}mez{-}Bombarelli}, \bibinfo{person}{Timothy Hirzel}, \bibinfo{person}{Al{\'{a}}n Aspuru{-}Guzik}, {and} \bibinfo{person}{Ryan~P. Adams}.} \bibinfo{year}{2015}\natexlab{}.
\newblock \showarticletitle{Convolutional Networks on Graphs for Learning Molecular Fingerprints}. In \bibinfo{booktitle}{\emph{Proc. {NeurIPS}}}. \bibinfo{pages}{2224--2232}.
\newblock


\bibitem[Gong et~al\mbox{.}(2021)]%
        {DBLP:journals/bdr/GongWWWL21}
\bibfield{author}{\bibinfo{person}{Fan Gong}, \bibinfo{person}{Meng Wang}, \bibinfo{person}{Haofen Wang}, \bibinfo{person}{Sen Wang}, {and} \bibinfo{person}{Mengyue Liu}.} \bibinfo{year}{2021}\natexlab{}.
\newblock \showarticletitle{{SMR:} Medical Knowledge Graph Embedding for Safe Medicine Recommendation}.
\newblock \bibinfo{journal}{\emph{Big Data Res.}}  \bibinfo{volume}{23} (\bibinfo{year}{2021}), \bibinfo{pages}{100174}.
\newblock


\bibitem[Gu et~al\mbox{.}(2022)]%
        {DBLP:journals/health/GuTCLULNGP22}
\bibfield{author}{\bibinfo{person}{Yu Gu}, \bibinfo{person}{Robert Tinn}, \bibinfo{person}{Hao Cheng}, \bibinfo{person}{Michael Lucas}, \bibinfo{person}{Naoto Usuyama}, \bibinfo{person}{Xiaodong Liu}, \bibinfo{person}{Tristan Naumann}, \bibinfo{person}{Jianfeng Gao}, {and} \bibinfo{person}{Hoifung Poon}.} \bibinfo{year}{2022}\natexlab{}.
\newblock \showarticletitle{Domain-Specific Language Model Pretraining for Biomedical Natural Language Processing}.
\newblock \bibinfo{journal}{\emph{{ACM} Trans. Comput. Heal.}} \bibinfo{volume}{3}, \bibinfo{number}{1} (\bibinfo{year}{2022}), \bibinfo{pages}{2:1--2:23}.
\newblock


\bibitem[Han et~al\mbox{.}(2022)]%
        {NEURIPS2022_3bdeb28a}
\bibfield{author}{\bibinfo{person}{Jiaqi Han}, \bibinfo{person}{Wenbing Huang}, \bibinfo{person}{Tingyang Xu}, {and} \bibinfo{person}{Yu Rong}.} \bibinfo{year}{2022}\natexlab{}.
\newblock \showarticletitle{Equivariant Graph Hierarchy-Based Neural Networks}. In \bibinfo{booktitle}{\emph{Proc. {NeurIPS}}}, Vol.~\bibinfo{volume}{35}. \bibinfo{publisher}{Curran Associates, Inc.}, \bibinfo{pages}{9176--9187}.
\newblock


\bibitem[Huang et~al\mbox{.}(2020)]%
        {huang2020caster}
\bibfield{author}{\bibinfo{person}{Kexin Huang}, \bibinfo{person}{Cao Xiao}, \bibinfo{person}{Trong Hoang}, \bibinfo{person}{Lucas Glass}, {and} \bibinfo{person}{Jimeng Sun}.} \bibinfo{year}{2020}\natexlab{}.
\newblock \showarticletitle{Caster: Predicting drug interactions with chemical substructure representation}. In \bibinfo{booktitle}{\emph{Proceedings of the AAAI conference on artificial intelligence}}, Vol.~\bibinfo{volume}{34}. \bibinfo{pages}{702--709}.
\newblock


\bibitem[Huang et~al\mbox{.}(2022)]%
        {huang2022equivariant}
\bibfield{author}{\bibinfo{person}{Wenbing Huang}, \bibinfo{person}{Jiaqi Han}, \bibinfo{person}{Yu Rong}, \bibinfo{person}{Tingyang Xu}, \bibinfo{person}{Fuchun Sun}, {and} \bibinfo{person}{Junzhou Huang}.} \bibinfo{year}{2022}\natexlab{}.
\newblock \showarticletitle{Equivariant Graph Mechanics Networks with Constraints}. In \bibinfo{booktitle}{\emph{Proc. {ICLR}}}.
\newblock


\bibitem[Jiao et~al\mbox{.}(2023)]%
        {jiao2023energy}
\bibfield{author}{\bibinfo{person}{Rui Jiao}, \bibinfo{person}{Jiaqi Han}, \bibinfo{person}{Wenbing Huang}, \bibinfo{person}{Yu Rong}, {and} \bibinfo{person}{Yang Liu}.} \bibinfo{year}{2023}\natexlab{}.
\newblock \showarticletitle{Energy-motivated equivariant pretraining for 3d molecular graphs}. In \bibinfo{booktitle}{\emph{Proceedings of the AAAI Conference on Artificial Intelligence}}, Vol.~\bibinfo{volume}{37}. \bibinfo{pages}{8096--8104}.
\newblock


\bibitem[Johnson et~al\mbox{.}(2016)]%
        {johnson2016mimic}
\bibfield{author}{\bibinfo{person}{Alistair~EW Johnson}, \bibinfo{person}{Tom~J Pollard}, \bibinfo{person}{Lu Shen}, \bibinfo{person}{Li-wei~H Lehman}, \bibinfo{person}{Mengling Feng}, \bibinfo{person}{Mohammad Ghassemi}, \bibinfo{person}{Benjamin Moody}, \bibinfo{person}{Peter Szolovits}, \bibinfo{person}{Leo Anthony~Celi}, {and} \bibinfo{person}{Roger~G Mark}.} \bibinfo{year}{2016}\natexlab{}.
\newblock \showarticletitle{MIMIC-III, a freely accessible critical care database}.
\newblock \bibinfo{journal}{\emph{Scientific data}} \bibinfo{volume}{3}, \bibinfo{number}{1} (\bibinfo{year}{2016}), \bibinfo{pages}{1--9}.
\newblock


\bibitem[Johnson et~al\mbox{.}(2018)]%
        {johnson2018mimic}
\bibfield{author}{\bibinfo{person}{Alistair~EW Johnson}, \bibinfo{person}{David~J Stone}, \bibinfo{person}{Leo~A Celi}, {and} \bibinfo{person}{Tom~J Pollard}.} \bibinfo{year}{2018}\natexlab{}.
\newblock \showarticletitle{The MIMIC Code Repository: enabling reproducibility in critical care research}.
\newblock \bibinfo{journal}{\emph{Journal of the American Medical Informatics Association}} \bibinfo{volume}{25}, \bibinfo{number}{1} (\bibinfo{year}{2018}), \bibinfo{pages}{32--39}.
\newblock


\bibitem[Kim et~al\mbox{.}(2021)]%
        {DBLP:journals/nar/00020CGHHLSTYZ021}
\bibfield{author}{\bibinfo{person}{Sunghwan Kim}, \bibinfo{person}{Jie Chen}, \bibinfo{person}{Tiejun Cheng}, \bibinfo{person}{Asta Gindulyte}, \bibinfo{person}{Jia He}, \bibinfo{person}{Siqian He}, \bibinfo{person}{Qingliang Li}, \bibinfo{person}{Benjamin~A. Shoemaker}, \bibinfo{person}{Paul~A. Thiessen}, \bibinfo{person}{Bo Yu}, \bibinfo{person}{Leonid Zaslavsky}, \bibinfo{person}{Jian Zhang}, {and} \bibinfo{person}{Evan Bolton}.} \bibinfo{year}{2021}\natexlab{}.
\newblock \showarticletitle{PubChem in 2021: new data content and improved web interfaces}.
\newblock \bibinfo{journal}{\emph{Nucleic Acids Res.}} \bibinfo{volume}{49}, \bibinfo{number}{Database-Issue} (\bibinfo{year}{2021}), \bibinfo{pages}{D1388--D1395}.
\newblock


\bibitem[Krenn et~al\mbox{.}(2020)]%
        {DBLP:journals/mlst/KrennHNFA20}
\bibfield{author}{\bibinfo{person}{Mario Krenn}, \bibinfo{person}{Florian H{\"{a}}se}, \bibinfo{person}{AkshatKumar Nigam}, \bibinfo{person}{Pascal Friederich}, {and} \bibinfo{person}{Al{\'{a}}n Aspuru{-}Guzik}.} \bibinfo{year}{2020}\natexlab{}.
\newblock \showarticletitle{Self-referencing embedded strings {(SELFIES):} {A} 100{\%} robust molecular string representation}.
\newblock \bibinfo{journal}{\emph{Mach. Learn. Sci. Technol.}} \bibinfo{volume}{1}, \bibinfo{number}{4} (\bibinfo{year}{2020}), \bibinfo{pages}{45024}.
\newblock


\bibitem[Le et~al\mbox{.}(2018)]%
        {DBLP:conf/kdd/Le0V18}
\bibfield{author}{\bibinfo{person}{Hung Le}, \bibinfo{person}{Truyen Tran}, {and} \bibinfo{person}{Svetha Venkatesh}.} \bibinfo{year}{2018}\natexlab{}.
\newblock \showarticletitle{Dual Memory Neural Computer for Asynchronous Two-view Sequential Learning}. In \bibinfo{booktitle}{\emph{Proceedings of the ACM SIGKDD Conference on Knowledge Discovery and Data Mining}}. \bibinfo{publisher}{{ACM}}, \bibinfo{pages}{1637--1645}.
\newblock


\bibitem[Lee and Lee(2020)]%
        {lee2020clinical}
\bibfield{author}{\bibinfo{person}{Cecilia~S Lee} {and} \bibinfo{person}{Aaron~Y Lee}.} \bibinfo{year}{2020}\natexlab{}.
\newblock \showarticletitle{Clinical applications of continual learning machine learning}.
\newblock \bibinfo{journal}{\emph{The Lancet Digital Health}} \bibinfo{volume}{2}, \bibinfo{number}{6} (\bibinfo{year}{2020}), \bibinfo{pages}{e279--e281}.
\newblock


\bibitem[Lee et~al\mbox{.}(2020)]%
        {DBLP:journals/bioinformatics/LeeYKKKSK20}
\bibfield{author}{\bibinfo{person}{Jinhyuk Lee}, \bibinfo{person}{Wonjin Yoon}, \bibinfo{person}{Sungdong Kim}, \bibinfo{person}{Donghyeon Kim}, \bibinfo{person}{Sunkyu Kim}, \bibinfo{person}{Chan~Ho So}, {and} \bibinfo{person}{Jaewoo Kang}.} \bibinfo{year}{2020}\natexlab{}.
\newblock \showarticletitle{BioBERT: a pre-trained biomedical language representation model for biomedical text mining}.
\newblock \bibinfo{journal}{\emph{Bioinform.}} \bibinfo{volume}{36}, \bibinfo{number}{4} (\bibinfo{year}{2020}), \bibinfo{pages}{1234--1240}.
\newblock


\bibitem[Li et~al\mbox{.}(2024)]%
        {li2024neural}
\bibfield{author}{\bibinfo{person}{Xuan Li}, \bibinfo{person}{Zhanke Zhou}, \bibinfo{person}{Jiangchao Yao}, \bibinfo{person}{Yu Rong}, \bibinfo{person}{Lu Zhang}, {and} \bibinfo{person}{Bo Han}.} \bibinfo{year}{2024}\natexlab{}.
\newblock \showarticletitle{Neural Atoms: Propagating Long-range Interaction in Molecular Graphs through Efficient Communication Channel}. In \bibinfo{booktitle}{\emph{Proc. {ICLR}}}.
\newblock


\bibitem[Liu et~al\mbox{.}(2019)]%
        {DBLP:conf/nips/LiuDL19}
\bibfield{author}{\bibinfo{person}{Shengchao Liu}, \bibinfo{person}{Mehmet~Furkan Demirel}, {and} \bibinfo{person}{Yingyu Liang}.} \bibinfo{year}{2019}\natexlab{}.
\newblock \showarticletitle{N-Gram Graph: Simple Unsupervised Representation for Graphs, with Applications to Molecules}. In \bibinfo{booktitle}{\emph{Proc. {NeurIPS}}}. \bibinfo{pages}{8464--8476}.
\newblock


\bibitem[Luo et~al\mbox{.}(2021)]%
        {luo2021ecnet}
\bibfield{author}{\bibinfo{person}{Yunan Luo}, \bibinfo{person}{Guangde Jiang}, \bibinfo{person}{Tianhao Yu}, \bibinfo{person}{Yang Liu}, \bibinfo{person}{Lam Vo}, \bibinfo{person}{Hantian Ding}, \bibinfo{person}{Yufeng Su}, \bibinfo{person}{Wesley~Wei Qian}, \bibinfo{person}{Huimin Zhao}, {and} \bibinfo{person}{Jian Peng}.} \bibinfo{year}{2021}\natexlab{}.
\newblock \showarticletitle{{ECNet} is an evolutionary context-integrated deep learning framework for protein engineering}.
\newblock \bibinfo{journal}{\emph{Nature Communications}} \bibinfo{volume}{12}, \bibinfo{number}{1} (\bibinfo{date}{Sept.} \bibinfo{year}{2021}).
\newblock


\bibitem[Ma et~al\mbox{.}(2022)]%
        {10.1093/bioinformatics/btac039}
\bibfield{author}{\bibinfo{person}{Hehuan Ma}, \bibinfo{person}{Yatao Bian}, \bibinfo{person}{Yu Rong}, \bibinfo{person}{Wenbing Huang}, \bibinfo{person}{Tingyang Xu}, \bibinfo{person}{Weiyang Xie}, \bibinfo{person}{Geyan Ye}, {and} \bibinfo{person}{Junzhou Huang}.} \bibinfo{year}{2022}\natexlab{}.
\newblock \showarticletitle{{Cross-dependent graph neural networks for molecular property prediction}}.
\newblock \bibinfo{journal}{\emph{Bioinform.}} \bibinfo{volume}{38}, \bibinfo{number}{7} (\bibinfo{date}{01} \bibinfo{year}{2022}), \bibinfo{pages}{2003--2009}.
\newblock
\showISSN{1367-4803}


\bibitem[Mu et~al\mbox{.}(2022)]%
        {mu2022learning}
\bibfield{author}{\bibinfo{person}{Zongshen Mu}, \bibinfo{person}{Yueting Zhuang}, \bibinfo{person}{Jie Tan}, \bibinfo{person}{Jun Xiao}, {and} \bibinfo{person}{Siliang Tang}.} \bibinfo{year}{2022}\natexlab{}.
\newblock \showarticletitle{Learning hybrid behavior patterns for multimedia recommendation}. In \bibinfo{booktitle}{\emph{Proceedings of the 30th ACM International Conference on Multimedia}}. \bibinfo{pages}{376--384}.
\newblock


\bibitem[Peng et~al\mbox{.}(2019)]%
        {DBLP:conf/bionlp/PengYL19}
\bibfield{author}{\bibinfo{person}{Yifan Peng}, \bibinfo{person}{Shankai Yan}, {and} \bibinfo{person}{Zhiyong Lu}.} \bibinfo{year}{2019}\natexlab{}.
\newblock \showarticletitle{Transfer Learning in Biomedical Natural Language Processing: An Evaluation of {BERT} and ELMo on Ten Benchmarking Datasets}. In \bibinfo{booktitle}{\emph{Proceedings of the 18th BioNLP Workshop and Shared Task, BioNLP@ACL}}, \bibfield{editor}{\bibinfo{person}{Dina Demner{-}Fushman}, \bibinfo{person}{Kevin~Bretonnel Cohen}, \bibinfo{person}{Sophia Ananiadou}, {and} \bibinfo{person}{Junichi Tsujii}} (Eds.). \bibinfo{publisher}{ACL}, \bibinfo{pages}{58--65}.
\newblock


\bibitem[Pollard et~al\mbox{.}(2018)]%
        {pollard2018eicu}
\bibfield{author}{\bibinfo{person}{Tom~J Pollard}, \bibinfo{person}{Alistair~EW Johnson}, \bibinfo{person}{Jesse~D Raffa}, \bibinfo{person}{Leo~A Celi}, \bibinfo{person}{Roger~G Mark}, {and} \bibinfo{person}{Omar Badawi}.} \bibinfo{year}{2018}\natexlab{}.
\newblock \showarticletitle{The eICU Collaborative Research Database, a freely available multi-center database for critical care research}.
\newblock \bibinfo{journal}{\emph{Scientific data}} \bibinfo{volume}{5}, \bibinfo{number}{1} (\bibinfo{year}{2018}), \bibinfo{pages}{1--13}.
\newblock


\bibitem[Read et~al\mbox{.}(2009)]%
        {DBLP:conf/pkdd/ReadPHF09}
\bibfield{author}{\bibinfo{person}{Jesse Read}, \bibinfo{person}{Bernhard Pfahringer}, \bibinfo{person}{Geoffrey Holmes}, {and} \bibinfo{person}{Eibe Frank}.} \bibinfo{year}{2009}\natexlab{}.
\newblock \showarticletitle{Classifier Chains for Multi-label Classification}. In \bibinfo{booktitle}{\emph{{ECML} {PKDD}}}. \bibinfo{publisher}{Springer}, \bibinfo{pages}{254--269}.
\newblock


\bibitem[Rong et~al\mbox{.}(2020)]%
        {NEURIPS2020_94aef384}
\bibfield{author}{\bibinfo{person}{Yu Rong}, \bibinfo{person}{Yatao Bian}, \bibinfo{person}{Tingyang Xu}, \bibinfo{person}{Weiyang Xie}, \bibinfo{person}{Ying WEI}, \bibinfo{person}{Wenbing Huang}, {and} \bibinfo{person}{Junzhou Huang}.} \bibinfo{year}{2020}\natexlab{}.
\newblock \showarticletitle{Self-Supervised Graph Transformer on Large-Scale Molecular Data}. In \bibinfo{booktitle}{\emph{Proc. {NeurIPS}}}, \bibfield{editor}{\bibinfo{person}{H.~Larochelle}, \bibinfo{person}{M.~Ranzato}, \bibinfo{person}{R.~Hadsell}, \bibinfo{person}{M.F. Balcan}, {and} \bibinfo{person}{H.~Lin}} (Eds.), Vol.~\bibinfo{volume}{33}. \bibinfo{publisher}{Curran Associates, Inc.}, \bibinfo{pages}{12559--12571}.
\newblock


\bibitem[Ruksakulpiwat et~al\mbox{.}(2023)]%
        {ruksakulpiwat2023does}
\bibfield{author}{\bibinfo{person}{Suebsarn Ruksakulpiwat}, \bibinfo{person}{Wendie Zhou}, \bibinfo{person}{Atsadaporn Niyomyart}, \bibinfo{person}{Tongyao Wang}, {and} \bibinfo{person}{Aaron Kudlowitz}.} \bibinfo{year}{2023}\natexlab{}.
\newblock \showarticletitle{How does the COVID-19 pandemic impact medication adherence of patients with chronic disease?: A systematic review}.
\newblock \bibinfo{journal}{\emph{Chronic illness}} \bibinfo{volume}{19}, \bibinfo{number}{3} (\bibinfo{year}{2023}), \bibinfo{pages}{495--513}.
\newblock


\bibitem[Satorras et~al\mbox{.}(2021)]%
        {DBLP:conf/icml/SatorrasHW21}
\bibfield{author}{\bibinfo{person}{Victor~Garcia Satorras}, \bibinfo{person}{Emiel Hoogeboom}, {and} \bibinfo{person}{Max Welling}.} \bibinfo{year}{2021}\natexlab{}.
\newblock \showarticletitle{E(n) Equivariant Graph Neural Networks}. In \bibinfo{booktitle}{\emph{Proceedings of the 38th International Conference on Machine Learning, {ICML} 2021, 18-24 July 2021, Virtual Event}} \emph{(\bibinfo{series}{Proceedings of Machine Learning Research}, Vol.~\bibinfo{volume}{139})}, \bibfield{editor}{\bibinfo{person}{Marina Meila} {and} \bibinfo{person}{Tong Zhang}} (Eds.). \bibinfo{publisher}{{PMLR}}, \bibinfo{pages}{9323--9332}.
\newblock


\bibitem[Shang et~al\mbox{.}(2019)]%
        {DBLP:conf/aaai/ShangXMLS19}
\bibfield{author}{\bibinfo{person}{Junyuan Shang}, \bibinfo{person}{Cao Xiao}, \bibinfo{person}{Tengfei Ma}, \bibinfo{person}{Hongyan Li}, {and} \bibinfo{person}{Jimeng Sun}.} \bibinfo{year}{2019}\natexlab{}.
\newblock \showarticletitle{GAMENet: Graph Augmented MEmory Networks for Recommending Medication Combination}. In \bibinfo{booktitle}{\emph{Proceedings of the AAAI Conference on Artificial Intelligence}}. \bibinfo{publisher}{{AAAI} Press}, \bibinfo{pages}{1126--1133}.
\newblock


\bibitem[Shoenbill et~al\mbox{.}(2023)]%
        {shoenbill2023artificial}
\bibfield{author}{\bibinfo{person}{Kimberly~A Shoenbill}, \bibinfo{person}{Suranga~N Kasturi}, {and} \bibinfo{person}{Eneida~A Mendonca}.} \bibinfo{year}{2023}\natexlab{}.
\newblock \showarticletitle{Artificial Intelligence, Machine Learning, and Natural Language Processing}.
\newblock In \bibinfo{booktitle}{\emph{Chronic Illness Care: Principles and Practice}}. \bibinfo{publisher}{Springer}, \bibinfo{pages}{469--479}.
\newblock


\bibitem[Sun et~al\mbox{.}(2022)]%
        {DBLP:conf/nips/Sun0LCW022}
\bibfield{author}{\bibinfo{person}{Hongda Sun}, \bibinfo{person}{Shufang Xie}, \bibinfo{person}{Shuqi Li}, \bibinfo{person}{Yuhan Chen}, \bibinfo{person}{Ji{-}Rong Wen}, {and} \bibinfo{person}{Rui Yan}.} \bibinfo{year}{2022}\natexlab{}.
\newblock \showarticletitle{Debiased, Longitudinal and Coordinated Drug Recommendation through Multi-Visit Clinic Records}. In \bibinfo{booktitle}{\emph{Proc. {NeurIPS}}}.
\newblock


\bibitem[Van~der Maaten and Hinton(2008)]%
        {van2008visualizing}
\bibfield{author}{\bibinfo{person}{Laurens Van~der Maaten} {and} \bibinfo{person}{Geoffrey Hinton}.} \bibinfo{year}{2008}\natexlab{}.
\newblock \showarticletitle{Visualizing data using t-SNE.}
\newblock \bibinfo{journal}{\emph{Journal of machine learning research}} \bibinfo{volume}{9}, \bibinfo{number}{11} (\bibinfo{year}{2008}).
\newblock


\bibitem[Wang et~al\mbox{.}(2018)]%
        {DBLP:conf/kdd/WangZHZ18}
\bibfield{author}{\bibinfo{person}{Lu Wang}, \bibinfo{person}{Wei Zhang}, \bibinfo{person}{Xiaofeng He}, {and} \bibinfo{person}{Hongyuan Zha}.} \bibinfo{year}{2018}\natexlab{}.
\newblock \showarticletitle{Supervised Reinforcement Learning with Recurrent Neural Network for Dynamic Treatment Recommendation}. In \bibinfo{booktitle}{\emph{Proceedings of the ACM SIGKDD Conference on Knowledge Discovery and Data Mining}}. \bibinfo{publisher}{{ACM}}, \bibinfo{pages}{2447--2456}.
\newblock


\bibitem[Wang et~al\mbox{.}(2019)]%
        {DBLP:conf/cikm/WangRCR0R19}
\bibfield{author}{\bibinfo{person}{Shanshan Wang}, \bibinfo{person}{Pengjie Ren}, \bibinfo{person}{Zhumin Chen}, \bibinfo{person}{Zhaochun Ren}, \bibinfo{person}{Jun Ma}, {and} \bibinfo{person}{Maarten de Rijke}.} \bibinfo{year}{2019}\natexlab{}.
\newblock \showarticletitle{Order-free Medicine Combination Prediction with Graph Convolutional Reinforcement Learning}. In \bibinfo{booktitle}{\emph{Proceedings of the ACM International Conference on Information and Knowledge Management}}. \bibinfo{publisher}{{ACM}}, \bibinfo{pages}{1623--1632}.
\newblock


\bibitem[Wishart et~al\mbox{.}(2018)]%
        {DBLP:journals/nar/WishartFGLMGSJL18}
\bibfield{author}{\bibinfo{person}{David~S. Wishart}, \bibinfo{person}{Yannick~D. Feunang}, \bibinfo{person}{An~Chi Guo}, \bibinfo{person}{Elvis~J. Lo}, \bibinfo{person}{Ana Marcu}, \bibinfo{person}{Jason~R. Grant}, \bibinfo{person}{Tanvir Sajed}, \bibinfo{person}{Daniel Johnson}, \bibinfo{person}{Carin Li}, \bibinfo{person}{Zinat Sayeeda}, \bibinfo{person}{Nazanin Assempour}, \bibinfo{person}{Ithayavani Iynkkaran}, \bibinfo{person}{Yifeng Liu}, \bibinfo{person}{Adam Maciejewski}, \bibinfo{person}{Nicola Gale}, \bibinfo{person}{Alex Wilson}, \bibinfo{person}{Lucy Chin}, \bibinfo{person}{Ryan Cummings}, \bibinfo{person}{Diana Le}, \bibinfo{person}{Allison Pon}, \bibinfo{person}{Craig Knox}, {and} \bibinfo{person}{Michael Wilson}.} \bibinfo{year}{2018}\natexlab{}.
\newblock \showarticletitle{DrugBank 5.0: a major update to the DrugBank database for 2018}.
\newblock \bibinfo{journal}{\emph{Nucleic Acids Res.}} \bibinfo{volume}{46}, \bibinfo{number}{Database-Issue} (\bibinfo{year}{2018}), \bibinfo{pages}{D1074--D1082}.
\newblock


\bibitem[Wu et~al\mbox{.}(2022)]%
        {DBLP:conf/www/WuQJQW22}
\bibfield{author}{\bibinfo{person}{Rui Wu}, \bibinfo{person}{Zhaopeng Qiu}, \bibinfo{person}{Jiacheng Jiang}, \bibinfo{person}{Guilin Qi}, {and} \bibinfo{person}{Xian Wu}.} \bibinfo{year}{2022}\natexlab{}.
\newblock \showarticletitle{Conditional Generation Net for Medication Recommendation}. In \bibinfo{booktitle}{\emph{Proceedings of the ACM Web Conference}}. \bibinfo{publisher}{{ACM}}, \bibinfo{pages}{935--945}.
\newblock


\bibitem[Yang et~al\mbox{.}(2021a)]%
        {DBLP:conf/ijcai/YangXGS21}
\bibfield{author}{\bibinfo{person}{Chaoqi Yang}, \bibinfo{person}{Cao Xiao}, \bibinfo{person}{Lucas Glass}, {and} \bibinfo{person}{Jimeng Sun}.} \bibinfo{year}{2021}\natexlab{a}.
\newblock \showarticletitle{Change Matters: Medication Change Prediction with Recurrent Residual Networks}. In \bibinfo{booktitle}{\emph{Proceedings of the International Joint Conference on Artificial Intelligence}}. \bibinfo{publisher}{ijcai.org}, \bibinfo{pages}{3728--3734}.
\newblock


\bibitem[Yang et~al\mbox{.}(2021b)]%
        {DBLP:conf/ijcai/YangXMGS21}
\bibfield{author}{\bibinfo{person}{Chaoqi Yang}, \bibinfo{person}{Cao Xiao}, \bibinfo{person}{Fenglong Ma}, \bibinfo{person}{Lucas Glass}, {and} \bibinfo{person}{Jimeng Sun}.} \bibinfo{year}{2021}\natexlab{b}.
\newblock \showarticletitle{SafeDrug: Dual Molecular Graph Encoders for Recommending Effective and Safe Drug Combinations}. In \bibinfo{booktitle}{\emph{Proceedings of the International Joint Conference on Artificial Intelligence}}, \bibfield{editor}{\bibinfo{person}{Zhi{-}Hua Zhou}} (Ed.). \bibinfo{pages}{3735--3741}.
\newblock


\bibitem[Yang et~al\mbox{.}(2023)]%
        {DBLP:conf/www/YangZWY23}
\bibfield{author}{\bibinfo{person}{Nianzu Yang}, \bibinfo{person}{Kaipeng Zeng}, \bibinfo{person}{Qitian Wu}, {and} \bibinfo{person}{Junchi Yan}.} \bibinfo{year}{2023}\natexlab{}.
\newblock \showarticletitle{MoleRec: Combinatorial Drug Recommendation with Substructure-Aware Molecular Representation Learning}. In \bibinfo{booktitle}{\emph{Proceedings of the ACM Web Conference}}. \bibinfo{publisher}{{ACM}}, \bibinfo{pages}{4075--4085}.
\newblock


\bibitem[Zhang et~al\mbox{.}(2017)]%
        {zhang2017leap}
\bibfield{author}{\bibinfo{person}{Yutao Zhang}, \bibinfo{person}{Robert Chen}, \bibinfo{person}{Jie Tang}, \bibinfo{person}{Walter~F Stewart}, {and} \bibinfo{person}{Jimeng Sun}.} \bibinfo{year}{2017}\natexlab{}.
\newblock \showarticletitle{LEAP: learning to prescribe effective and safe treatment combinations for multimorbidity}. In \bibinfo{booktitle}{\emph{Proceedings of the ACM SIGKDD Conference on Knowledge Discovery and Data Mining}}. \bibinfo{pages}{1315--1324}.
\newblock


\bibitem[Zhou et~al\mbox{.}(2023)]%
        {DBLP:journals/corr/abs-2302-04473}
\bibfield{author}{\bibinfo{person}{Hongyu Zhou}, \bibinfo{person}{Xin Zhou}, \bibinfo{person}{Zhiwei Zeng}, \bibinfo{person}{Lingzi Zhang}, {and} \bibinfo{person}{Zhiqi Shen}.} \bibinfo{year}{2023}\natexlab{}.
\newblock \showarticletitle{A Comprehensive Survey on Multimodal Recommender Systems: Taxonomy, Evaluation, and Future Directions}.
\newblock \bibinfo{journal}{\emph{CoRR}}  \bibinfo{volume}{abs/2302.04473} (\bibinfo{year}{2023}).
\newblock


\end{thebibliography}

\end{document}